\pdfoutput=1
\documentclass[acmsmall,screen]{acmart}

\AtBeginDocument{%
  }

\setcopyright{acmlicensed}
\copyrightyear{2025}
\acmYear{2025}
\acmDOI{XXXXXXX.XXXXXXX}

\acmJournal{JACM}
\acmVolume{37}
\acmNumber{4}
\acmArticle{111}
\acmMonth{8}

\usepackage{subfigure}
\usepackage{amsmath,amsfonts}
\usepackage{array}
\usepackage{textcomp}
\usepackage{stfloats}
\usepackage{url}
\usepackage{wrapfig}
\usepackage{verbatim}
\usepackage{graphicx}
\usepackage{lipsum}
\usepackage{tabularx}
\usepackage{booktabs}
\usepackage{multirow}
\usepackage{makecell}
\usepackage{algorithm}
\usepackage{algorithmic}
\usepackage{colortbl}
\usepackage{threeparttable}
\usepackage{xcolor}
\definecolor{mygreen}{rgb}{0.7, 1.0, 0.7}
\definecolor{myblue}{rgb}{0.7, 0.8, 1.0}
\newcommand{\percentbarri}[4]{%
  \rlap{\color{#3}\hspace*{-3pt}\rule{#2 em}{1.6ex}}%
  \makebox[#4 em][r]{#1}%
}

\begin{document}
\title{InterCLIP-MEP: Interactive CLIP and Memory-Enhanced Predictor for Multi-modal Sarcasm Detection}

\author{Junjie Chen}
\affiliation{%
  \institution{Anhui Polytechnic University}
  \city{Wuhu}
  \country{China}
}
\email{chenjunjie@stu.ahpu.edu.cn}
\orcid{0009-0001-5288-048X}

\author{Hang Yu}
\affiliation{%
  \institution{Shanghai University}
  \city{Shanghai}
  \country{China}
}
\email{yuhang@shu.edu.cn}
\authornote{Co-corresponding author: Hang Yu.}
\orcid{0000-0003-3444-9992}

\author{Subin Huang}
\affiliation{%
  \institution{Anhui Polytechnic University}
  \city{Wuhu}
  \country{China}
}
\email{subinhuang@ahpu.edu.cn}
\authornote{Corresponding author: Subin Huang.}
\orcid{0000-0003-1886-4192}

\author{Sanmin Liu}
\affiliation{%
  \institution{Anhui Polytechnic University}
  \city{Wuhu}
  \country{China}
}
\email{sanmin.liu@ahpu.edu.cn}
\orcid{0000-0002-0399-7737}

\author{Linfeng Zhang}
\affiliation{%
  \institution{Shanghai Jiao Tong University}
  \city{Shanghai}
  \country{China}
}
\email{zhanglinfeng@sjtu.edu.cn}
\orcid{0000-0002-3341-183X}
\renewcommand{\shortauthors}{Junjie Chen, Hang Yu, Subin Huang, et al.}
\begin{abstract}
Sarcasm in social media, frequently conveyed through the interplay of text and images, presents significant challenges for sentiment analysis and intention mining.
Existing multi-modal sarcasm detection approaches have been shown to excessively depend on superficial cues within the textual modality, exhibiting limited capability to accurately discern sarcasm through subtle text-image interactions.
To address this limitation, a novel framework, InterCLIP-MEP, is proposed.
This framework integrates Interactive CLIP (InterCLIP), which employs an efficient training strategy to derive enriched cross-modal representations by embedding inter-modal information directly into each encoder, while using approximately 20.6$\times$ fewer trainable parameters compared with existing state-of-the-art (SOTA) methods.
Furthermore, a Memory-Enhanced Predictor (MEP) is introduced, featuring a dynamic dual-channel memory mechanism that captures and retains valuable knowledge from test samples during inference, serving as a non-parametric classifier to enhance sarcasm detection robustness.
Extensive experiments on MMSD, MMSD2.0, and DocMSU show that InterCLIP-MEP achieves SOTA performance, specifically improving accuracy by 1.08\% and F1 score by 1.51\% on MMSD2.0.
Under distributional shift evaluation, it attains 73.96\% accuracy, exceeding its memory-free variant by nearly 10\% and the previous SOTA by over 15\%, demonstrating superior stability and adaptability.
The implementation of InterCLIP-MEP is publicly available at \url{https://github.com/CoderChen01/InterCLIP-MEP}.
\end{abstract}

\received{30 April 2025}
\received[revised]{13 July 2025}
\received[revised]{25 October 2025}
\received[accepted]{05 November 2025}
\maketitle

\section{Introduction}

Sarcasm, characterized by its nuanced and intricate nature, serves as a vital mechanism in communication, often employed to convey irony, mockery, or implicit meanings~\cite{Gibbs1991PsychologicalAO,tsur2010icwsm,JingSOJN23,LiuZS24}.
The automated identification of sarcasm in textual data has evolved into a significant research focus, offering substantial support for tasks such as sentiment analysis and intent mining~\cite{pang2008opinion, JoshiTPBC16,JoshiBC17}.
With the growing prevalence of social media platforms such as Twitter and Reddit, users increasingly rely on the fusion of textual and visual elements to articulate their messages.
Consequently, multi-modal sarcasm detection has gained considerable prominence, presenting challenges in deciphering the intricate interplay between textual and visual cues to accurately identify sarcastic content~\cite{WangYJMXL24,ZhuangLZGZM25}.

Despite significant progress achieved by prior studies~\cite{xu2020reasoning, pan-etal-2020-modeling, liang2021multi, Liang2022MultiModalSD,10.24963/ijcai.2024/887} in this domain, Qin et al.~\cite{qin2023mmsd2} have identified the presence of spurious cues within the MMSD benchmark~\cite{Cai2019MultiModalSD} utilized in these works.
These cues contribute to the development of biased models, thereby overestimating their effectiveness in detecting sarcasm cues from multi-modal data.
Through the introduction of MMSD2.0, a refined benchmark that eliminates spurious cues and rectifies mislabeled samples, Qin et al.~\cite{qin2023mmsd2} reveal a significant performance degradation of existing methods when evaluated on MMSD2.0.
Although their proposed method, which leverages Multi-view CLIP for reliable multi-modal sarcasm detection, shows promise, it still exhibits limited capability in accurately discerning sarcasm through subtle text-image interactions.

\begin{wrapfigure}{r}{0.54\textwidth}
\begin{center}
\includegraphics[width=\linewidth]{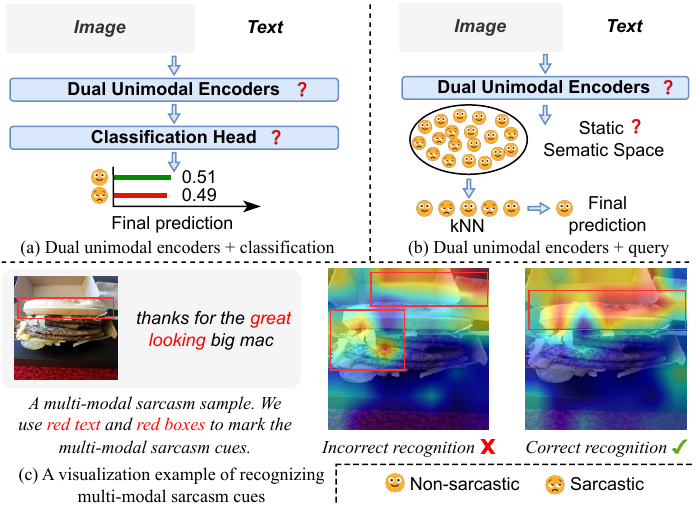}
\end{center}
\caption{
An overview of the limitations inherent in existing multi-modal sarcasm detection frameworks.
Panels (a) and (b) illustrate two predominant multi-modal sarcasm detection pipelines, with their respective shortcomings highlighted by \textit{red question marks}.
Panel (c) provides a visual representation of multi-modal sarcasm cues, demonstrating instances where such cues are either accurately identified or misinterpreted in a multi-modal sarcasm sample.
}
\label{fig:motivation}
\end{wrapfigure}

As depicted in Figures~\ref{fig:motivation}(a) and~\ref{fig:motivation}(b), numerous approaches utilize dual unimodal pre-trained encoders, such as ViT~\cite{dosovitskiy2020image} and BERT~\cite{devlin2019bert}, as the backbone for encoding text-image pairs, followed by specialized feature fusion techniques~\cite{xu2020reasoning, pan-etal-2020-modeling, liang2021multi, Liang2022MultiModalSD, wen2023dip, Tian2023DynamicRT, wei2024g2sam, Wang2025AdaSFFuse}.
However, such methodologies may fail to effectively capture multi-modal sarcasm cues compared to multi-modal pre-trained models like CLIP~\cite{radford2021learning}.
In Figure~\ref{fig:motivation}(a), the adoption of a learnable classification head to predict labels from fused representations is prevalent but often accompanied by high predictive entropy and substantial uncertainty.
The work by~\cite{wei2024g2sam} introduced the construction of a static semantic space utilizing historical training samples, where robust predictions during inference are achieved through KNN-based querying and voting, as illustrated in Figure~\ref{fig:motivation}(b).
Nevertheless, despite CLIP's demonstrated effectiveness as a text-image encoder for multi-modal sarcasm detection~\cite{qin2023mmsd2}, it remains challenged in capturing multi-modal sarcasm cues due to the inconsistency of sarcasm, which contradicts CLIP's alignment of text and image.
Additionally, reliance on a static semantic space for inference is inadequate to address the dynamic nature of evolving sample distributions.
Indeed, Qin et al.~\cite{qin2023mmsd2} have highlighted that many models depend on spurious cues within the MMSD benchmark~\cite{Cai2019MultiModalSD}, resulting in biased outcomes.

In this paper, building on these insights, we propose InterCLIP-MEP, a novel framework designed to enhance the reliability of multi-modal sarcasm detection.
Our InterCLIP-MEP introduces a refined variant of CLIP, Interactive CLIP (InterCLIP), as the backbone to enhance sample encoding for detecting multi-modal sarcasm.
InterCLIP directly embeds cross-modal representations into text and vision encoders, thereby enhancing the understanding of multi-modal sarcasm cues (as illustrated in Figure~\ref{fig:framework}, on the left).
To complement this, we further design a Memory-Enhanced Predictor (MEP), which dynamically leverages historical test sample features to construct a more adaptive and reliable non-parametric classifier (Figure~\ref{fig:framework}, right).
Importantly, the design of MEP is conceptually motivated by both cognitive science and machine learning theories.
From the cognitive perspective, human working memory is known to have limited capacity and tends to preferentially retain information with higher certainty~\cite{Baddeley2000TheEB,anderson1991reflections}. Inspired by this mechanism, MEP employs an entropy-based update strategy to selectively preserve low-uncertainty features, simulating human-like selective retention.
From the machine learning perspective, entropy is a widely recognized measure of predictive uncertainty, and uncertainty-aware sample selection has been shown effective in active learning and memory-based inference~\cite{toneva2018empirical,settles2009active}.
Beyond these general motivations, the characteristics of sarcasm further justify our design: sarcastic expressions in social media typically fall into a limited set of semantic patterns (e.g., exaggeration, contrast, or ironic praise), and humans often rely on prior experiences or historical contexts to correctly interpret them~\cite{attardo2000irony,gibbs1986psycholinguistics,gibbs2007irony,bamman2015contextualized,wallace-etal-2014-humans}.
By maintaining low-entropy representations in memory, MEP effectively acts as a denoising mechanism, ensuring more robust predictions during inference.
Collectively, these components constitute the proposed framework, InterCLIP-MEP.
Additionally, InterCLIP-MEP incorporates an efficient training strategy that fine-tunes cross-modal interactions through a lightweight adaptation mechanism, ensuring computational efficiency while achieving state-of-the-art performance (Figure~\ref{fig:framework}, left).

In summary, the key contributions of this work are as follows:
\begin{itemize}
\item The proposed framework, InterCLIP-MEP, introduces a novel approach for multi-modal sarcasm detection.
It integrates Interactive CLIP (InterCLIP) to enhance text-image interaction encoding and incorporates a Memory-Enhanced Predictor (MEP) to achieve more robust and reliable sarcasm predictions.
\item An efficient training strategy is proposed, which substantially reduces computational overhead compared to existing state-of-the-art methods.
By introducing approximately 20.6$\times$ fewer trainable parameters, the approach decreases GPU memory usage by about 2.5$\times$ and accelerates computation by roughly 8.7$\times$ with a batch size of 128, while maintaining superior performance on a single NVIDIA RTX 4090 GPU.
\item Extensive experiments conducted on the MMSD, MMSD2.0 and DocMSU benchmarks demonstrate that InterCLIP-MEP achieves significant improvements, outperforming state-of-the-art methods with an increase of 1.08\% in accuracy and 1.51\% in F1 score, particularly on the MMSD2.0 dataset.
\end{itemize}

The remainder of this paper is structured as follows.
Section~\ref{sec:related_work} reviews related studies and discusses the limitations of existing methodologies.
Section~\ref{sec:methodology} introduces the proposed InterCLIP-MEP framework, including the Interactive CLIP backbone, efficient training strategy, and Memory-Enhanced Predictor.
Section~\ref{sec:experimentsI} presents the main experiments, which provide both quantitative and qualitative evaluations as well as an efficiency analysis to demonstrate the effectiveness of the proposed method.
Section~\ref{sec:experimentsII} reports the extended experiments, covering four aspects: performance on a document-level benchmark, evaluation under low-resource settings, error propagation robustness study, and robustness analysis under distributional shifts, to comprehensively verify the generalization and stability of InterCLIP-MEP.
Section~\ref{sec:limitation-and-future-works} discusses the limitations of the current study and outlines potential directions for future research.
Finally, Section~\ref{sec:conclusion} concludes the paper.

\section{Related Work}
\label{sec:related_work}
\subsection{Multi-modal Sarcasm Detection}
Early research in sarcasm detection predominantly focused on textual data~\cite{bouazizi2015sarcasm,amir2016modelling,baziotis2018ntua}.
With the proliferation of social media platforms, identifying sarcasm in text-image pairs has emerged as a more complex challenge, prompting the advancement of multi-modal methodologies.
Among the pioneering efforts, Schifanella et al.~\cite{schifanella2016detecting} explored multi-modal social media posts to detect sarcasm cues.
Expanding on this work, Cai et al.~\cite{Cai2019MultiModalSD} proposed the MMSD benchmark, showcasing the efficacy of a hierarchical fusion model that incorporates image features.
This benchmark has since served as a cornerstone for multi-modal sarcasm detection, motivating a series of subsequent studies~\cite{xu2020reasoning,pan-etal-2020-modeling,liang2021multi,Liang2022MultiModalSD,Liu2022TowardsMS,qin2023mmsd2,wen2023dip,Tian2023DynamicRT,wei2024g2sam,10.24963/ijcai.2024/887,ding2022multi,10477507,zhuang2025multi}.
However, subsequent studies revealed that the MMSD benchmark contained spurious cues, which could introduce bias into model predictions~\cite{qin2023mmsd2}.
To address this issue, Qin et al.~\cite{qin2023mmsd2} developed the MMSD2.0 benchmark, which eliminates these cues and rectifies mislabeled samples.
Re-evaluations conducted on MMSD2.0 demonstrated significant performance declines in existing models, underscoring the necessity for more robust methodologies.

Concurrently, Tang et al.~\cite{tang-etal-2024-leveraging} explored the use of large vision-language models (LVLMs), such as instruction-tuned LLaVA variants, for multi-modal sarcasm detection by integrating prompting strategies and retrieval modules. While LVLMs demonstrate impressive general capabilities across a wide range of multi-modal tasks, their application to sarcasm detection yields only modest performance improvements.
More critically, these models involve billions of parameters, resulting in high computational costs and memory demands, which make them impractical for lightweight and scalable sarcasm detection.
Similar to prior sarcasm detection studies, which primarily compare against task-specific multi-modal architectures rather than general-purpose LVLMs, we also do not benchmark against LVLMs.
Instead, motivated by their limitations, we propose InterCLIP-MEP, which achieves state-of-the-art accuracy while requiring over 20$\times$ fewer trainable parameters and 2.5$\times$ lower GPU memory.
By addressing the shortcomings of both traditional fusion methods and LVLM-based approaches, our framework offers a practical and scalable solution for multi-modal sarcasm detection.

\subsection{CLIP Adaptation}
The Contrastive Language-Image Pretraining (CLIP) model~\cite{radford2021learning} exhibits remarkable capabilities in cross-modal alignment for vision-language understanding tasks. Recent progress in domain-specific adaptations of CLIP has led to substantial performance enhancements across diverse applications, including phrase localization~\cite{li2022adapting}, open-vocabulary semantic segmentation~\cite{Liang2023Open}, and action recognition~\cite{wang2023seeing}.
These achievements highlight CLIP's adaptability in addressing specialized visual-linguistic challenges.
Closely related to our work, Ganz et al.~\cite{ganz2024question} proposed the Question-Aware Vision Transformer (QA-ViT), which augments frozen vision encoders with question embeddings to produce question-aware visual representations.
However, QA-ViT was designed for general multi-modal reasoning tasks such as visual question answering rather than sarcasm detection.
It only explores the text-to-vision direction, leaving the reverse integration of visual cues into textual encoding unexplored.
More importantly, it does not address the pragmatic and incongruity-based nature of sarcasm, where subtle mismatches between modalities must be captured.

In parallel, Qin et al.~\cite{qin2023mmsd2} introduced Multi-view CLIP as the first CLIP-based approach for multi-modal sarcasm detection, where frozen CLIP encoders are used to extract text-only, image-only, and joint representations, and the outputs are fused via a transformer.
While pioneering, this design treats CLIP merely as a static feature extractor and lacks adaptive cross-modal interaction within the encoders.
In contrast, our Interactive CLIP (InterCLIP) explicitly conditions both encoders on cross-modal signals through bidirectional interaction and lightweight LoRA tuning, enabling a richer alignment of subtle sarcastic cues.
When combined with our Memory-Enhanced Predictor (MEP), which adaptively incorporates reliable exemplars at inference time, our framework directly targets the unique challenges of sarcasm detection and provides a principled, task-specific advancement over both QA-ViT and Multi-view CLIP.

\subsection{Memory-enhanced Prediction}
Inspired by the insights from cognitive science~\cite{stokes2015activity,Baddeley2000TheEB,anderson1991reflections,gibbs2007irony}, the concept of memory has been ingeniously integrated into neural networks to enhance their capabilities~\cite{weston2014memory,sukhbaatar2015end}.
This integration has opened up new avenues for improving model training, as evidenced by several pioneering studies~\cite{wu2018unsupervised,wen2023dip,zhu2020label}.
For instance, Wen et al.~\cite{wen2023dip} proposed DIP, which maintains semantic distributions in memory banks to stabilize training and better model incongruity, while Zhu and Yang~\cite{zhu2020label} introduced a label-independent memory (LIM) to cache unlabeled video features for robust prototype construction in semi-supervised few-shot learning.
Despite differences in domains, both DIP and LIM adopt training-time memory to act as stabilizers or prototype aggregators, and their memories vanish once the model enters the inference stage.
In contrast, our Memory-Enhanced Predictor (MEP) introduces a fundamentally different usage of memory: it is a test-time, entropy-aware predictor rather than a training aid.
Instead of serving as a regularizer during optimization, MEP dynamically maintains a compact buffer of reliable test exemplars and integrates them into the prediction process through cosine matching.
This makes memory a core inference component rather than an auxiliary training tool.

Furthermore, some recent works~\cite{zhang2024dual,wei2024g2sam} have also explored inference-time memories.
Zhang et al.~\cite{zhang2024dual} proposed Dual Memory Networks for general vision-language adaptation, where a static memory caches training features and a dynamic memory accumulates test representations, with cross-attention used to derive adaptive classifiers.
Wei et al.~\cite{wei2024g2sam} introduced G$^{2}$SAM for multi-modal sarcasm detection, which leverages a graph-based memory to capture global semantic relations.
However, both approaches rely on static or pre-constructed memory structures that do not evolve with incoming test samples.
By contrast, our MEP employs a dynamic, entropy-aware buffer that continuously updates with reliable test exemplars, making memory an adaptive and lightweight inference component tailored specifically for the challenges of sarcasm detection~\cite{gibbs2007irony,bamman2015contextualized}.

\begin{table}[ht]
\centering
\caption{
List of symbols.
}
\label{tab:symbol-list}
\begin{tabularx}{\textwidth}{lX}
\toprule
Symbol                                    & Description   \\
\midrule
$T$
& $T$ denotes a short text. \\

$I$
& $I$ represents an image. \\

$\mathcal{P}$                    
& $\mathcal{P}$ denotes a text-image pair $(T, I)$.  \\

$\mathcal{T}$
& $\mathcal{T}$ denotes CLIP's text encoder. \\

$\mathcal{V}$
& $\mathcal{V}$ denotes CLIP's vision encoder.\\

$\mathbf{F}$
& $\mathbf{F}$ represents the final layer representations encoded by either the text or vision encoder, with text representations as $\mathbf{F}_t$ and image representations as $\mathbf{F}_v$. \\

$\tilde{\mathbf{F}}$
& $\tilde{\mathbf{F}}$ represents the final layer representations encoded by either the text or vision encoder after embedding representations from another modality, with text representations as $\tilde{\mathbf{F}}_t$ and image representations as $\tilde{\mathbf{F}}_v$. \\

$\mathbf{H}$
& $\mathbf{H}$ represents the input representations for each sub-attention layer in the text or vision encoders. Each layer's input comes from the output of the previous layer, denoted $\mathbf{H}_t$ for the text encoder and $\mathbf{H}_v$ for the vision encoder.\\

$\mathcal{F}_{t/v}$
& $\mathcal{F}_{t/v}$ denotes the adapting projection layer in the text or vision encoders used to project the embedded representations of the other modality into the current encoder space. It is denoted as $\mathcal{F}_t$ in the text encoder and $\mathcal{F}_v$ in the vision encoder. \\

$\mathbf{F}^{'}$
&  $\mathbf{F}'$ represents the representations projected into the corresponding encoder space. For example, embedding visual representations $\mathbf{F}_v$ in the text encoder and projecting it through $\mathcal{F}_t$ results in $\mathbf{F}'_v$. \\

$\mathbf{H}^{'}$
&  $\mathbf{H}'$ represents the representations after embedding another modality's representations and processing them through a self-attention layer. \\

$\mathcal{H}_{t/v}$
&  $\mathcal{H}_{t/v}$ denotes the projection module in the self-attention layer used to transform the output of the self-attention module, denoted $\mathcal{H}_t$ for the text encoder and $\mathcal{H}_v$ for the vision encoder. \\

$\mathcal{G}_{t/v}$
& $\mathcal{G}_{t/v}$ denotes the projection module in the self-attention layer that has embedded representations from another modality, used to jointly transform the output representation in combination with $\mathcal{H}_{t/v}$. \\

$\mathbf{H}^{''}$
& $\mathbf{H}''$ represents the final representations in the self-attention layer. \\

$\tilde{h}^{f}$
&    $\tilde{h}^f$ denotes the final fused feature obtained from a sample. \\

$\mathcal{F}_{c}$
&  $\mathcal{F}_c$ denotes the classification module used to assign pseudo-labels to samples. \\

$\mathcal{F}_{p}$
& $\mathcal{F}_p$ denotes the projection module used to project samples into a latent space. \\

$\hat{h}^f$
&  $\hat{h}^f$ represents the feature of a sample's fused feature after transformation by $\mathcal{F}_p$ and L2 normalization. \\
\bottomrule
\end{tabularx}
\vspace{-1em}
\end{table}

\section{Methodology}
\label{sec:methodology}
\begin{figure*}[t]
\centering
\includegraphics[width=0.95\textwidth]{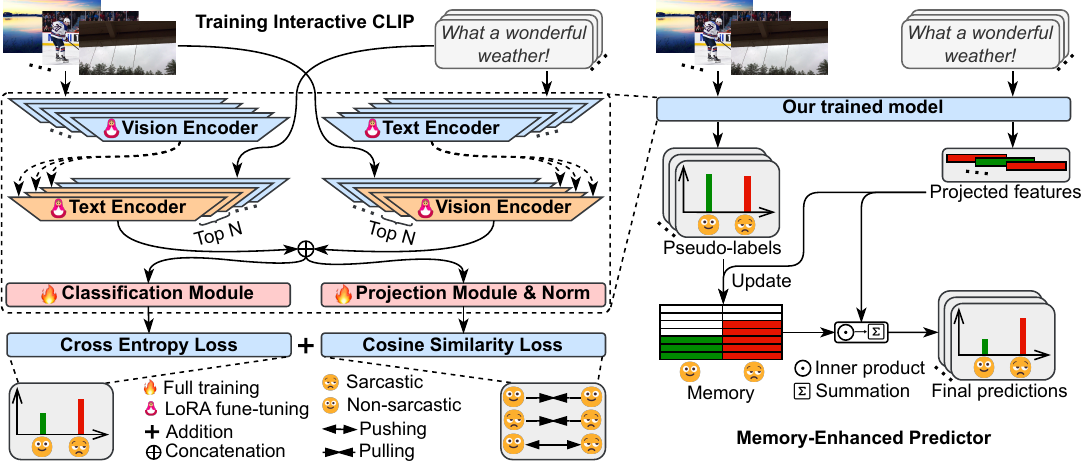}
\caption{
Overview of our framework.
\textbf{(I) Training Interactive CLIP (InterCLIP): }
Vision and text representations are extracted using separate encoders and embedded into the top-$n$ layers of the opposite modality's encoder for interaction.
The top-$n$ layers are fine-tuned with LoRA, while the rest of the encoder remains frozen.
Final vision and text representations are concatenated and used to train a classification module for identifying multi-modal sarcasm.
A projection module is also trained to project representations into a latent space.
\textbf{(II) Memory-Enhanced Predictor (MEP): }
During inference, InterCLIP generates interactive representations.
The classification module assigns pseudo-labels, and the projection module provides projection features.
MEP updates dynamic dual-channel memory with these features and pseudo-labels.
The final prediction of the current sample is made by comparing its projected feature with those in memory.
}
\label{fig:framework}
\vspace{-1em}
\end{figure*}

An overview of InterCLIP-MEP is illustrated in Figure~\ref{fig:framework}. 
The framework is first introduced through a detailed exposition of Interactive CLIP (InterCLIP) and its associated efficient training strategy, followed by a comprehensive analysis of the Memory-Enhanced Predictor (MEP).
For clarity, Table~\ref{tab:symbol-list} provides a summary of the primary mathematical symbols employed throughout this work.

\subsection{Interactive CLIP}
The input to Interactive CLIP (InterCLIP) is a text-image pair $\mathcal{P}=(T, I)$, where $T$ represents a piece of text and $I$ represents an image.
Here, for simplicity, we do not consider the case of batch inputs.
The text encoder $\mathcal{T}$ extracts the vanilla text representations $\mathbf{F}_{t}$:
\begin{equation}
\mathbf{F}_{t} = \mathcal{T}(T) = \{h^{t}_{\textrm{bos}}(t_{\textrm{bos}}), h^{t}_{1}(t_1), \dots, h^{t}_{n}(t_n), h^{t}_{\textrm{eos}}(t_{\textrm{eos}})\},
\end{equation}
where $t_i$ denotes a text token, $n$ is the length of $T$ after tokenization, $t_{\textrm{bos}}$ and $t_{\textrm{eos}}$ are special tokens required by the text encoder. 
Here, $h^{t}_{i}(\cdot) \in \mathbb{R}^{d_{t}}$ represents the $d_{t}$-dimensional encoded representation of the corresponding token $t_i$, with $i$ ranging from 1 to $n$, including the beginning-of-sequence (bos) and end-of-sequence (eos) tokens.

The vision encoder $\mathcal{V}$ extracts the vanilla image representations $\mathbf{F}_{v}$:
\begin{equation}
\mathbf{F}_{v}=\mathcal{V}(I)=\{h^{v}_{\textrm{cls}}(p_{\textrm{cls}}),h^{v}_{1}(p_1),\dots,h^{v}_{m}(p_m)\},
\end{equation}
where $I$ is processed into multiple patches $p_i$, $m$ is the number of patches, and $p_{\textrm{cls}}$ is a special token required by the visual encoder. 
Here, $h^{v}_{i}(\cdot)\in\mathbb{R}^{d_{v}}$ represents the $d_{v}$-dimensional encoded representation of the corresponding $p_i$, with $i$ ranging from 1 to $m$, including the classification (cls) token.
Specifically, both $\mathbf{F}_{t}$ and $\mathbf{F}_{v}$ are representations from the final layer outputs of their respective encoders.
Conditioning on $\mathbf{F}_{t}$ or $\mathbf{F}_{v}$, we can obtain the interactive text representations $\tilde{\mathbf{F}}_{t}$ or the interactive image representations $\tilde{\mathbf{F}}_{v}$:
\begin{equation}
\tilde{\mathbf{F}}_{t}=\mathcal{T}(T|\mathbf{F}_{v}),  \tilde{\mathbf{F}}_{v}=\mathcal{V}(V|\mathbf{F}_{t}).
\end{equation}

\begin{wrapfigure}{r}{0.5\textwidth}
\begin{center}
\includegraphics[width=\linewidth]{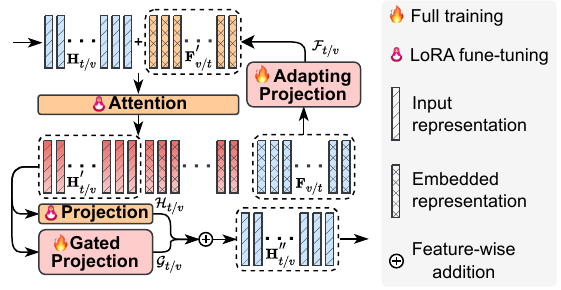}
\end{center}
\caption{
Structure of the conditional self-attention.
}
\label{fig:encoder-detail}
\end{wrapfigure}

We use $\tilde{h}^{t}_{i}(\cdot)\in\mathbb{R}^{d_{t}}$ and $\tilde{h}^{v}_{i}(\cdot)\in\mathbb{R}^{d_{v}}$ to denote the re-encoded interactive representations of each text token and image patch, respectively.

To be specific, we condition only the top-$n$ self-attention layers of the text or vision encoder, where $n$ is a hyperparameter that will be analyzed in the experiment section.
Figure~\ref{fig:encoder-detail} illustrates the structure of the conditioned self-attention layers.
Given that the text and vision encoder in CLIP share a similar architecture, for brevity, we denote the input representations to the self-attention layers of the text or vision encoder as $\mathbf{H}_{t/v}$, which are derived from the outputs of the previous layer.
The previous layer can either be conditioned or non-conditioned.
Due to the dimensional mismatch between the embedded representations $\mathbf{F}_{v/t}$ and the corresponding encoder representation space, we introduce an adapting projection layer $\mathcal{F}_{t/v}$ to project $\mathbf{F}_{v/t}$ into the appropriate representation space.

To fuse the input representations $\mathbf{H}_{t/v}$ with the projected embedded representations $\mathbf{F}_{v/t}^{'}=\mathcal{F}_{t/v}(\mathbf{F}_{v/t})$, we concatenate them and feed them into the attention layer to obtain the transformed representations.
We then extract the transformed input representations $\mathbf{H}^{'}_{t/v}$ from the output.
Following~\cite{ganz2024question}, we apply a gated projection layer $\mathcal{G}_{t/v}$ along with the self-attention's projection head $\mathcal{H}_{t/v}$ using a learnable gating mechanism to compute the self-attention output representations $\mathbf{H}^{''}_{t/v}$.
Given the similarity between the self-attention layers of the vision encoder and the text encoder, we use the text encoder $\mathcal{T}$ to illustrate the process as follows:
\begin{equation}
\mathbf{F}_{v}^{'} = \mathcal{F}_{t}(\mathbf{F}_{v}), \quad \mathbf{F}_{v} \in \mathbb{R}^{m \times d_{v}},\mathbf{F}_{v}^{'} \in \mathbb{R}^{m \times d_{t}},
\end{equation}
\begin{equation}
\mathbf{H}^{'}_{t} = \textrm{Attn}_{t}(\mathbf{H}_{t} \oplus \mathbf{F}_{v}^{'})_{[:n]}, \quad \mathbf{H}_{t}, \mathbf{H}^{'}_{t} \in \mathbb{R}^{n \times d_t}, \mathbf{H}_{t} \oplus \mathbf{F}_{v}^{'} \in \mathbb{R}^{(n+m) \times d_{t}},
\end{equation}
\begin{equation}
\mathbf{H}^{''}_{t} = \mathcal{H}_{t}(\mathbf{H}^{'}_{t}) + \mathcal{G}_{t}(\mathbf{H}^{'}_{t}) \cdot \textrm{tanh}(\beta_{t}),\mathbf{H}^{''}_{t} \in \mathbb{R}^{n \times d_{t}}.
\end{equation}

Here, $\oplus$ denotes the concatenation operation, and $\beta_{t}$ is a learnable gating parameter initialized to 0 to ensure training stability.
The subsequent computation follows the original CLIP, ultimately yielding the interactive representations $\tilde{\mathbf{F}}_{t}$.

InterCLIP supports three interaction modes for fusing text and image features into the final representation \(\tilde{h}^{f}\in\mathbb{R}^{d_{t}+d_{v}}\):  
\begin{itemize}
\item \textbf{T2V:} Text representations \(\mathbf{F}_{t}\) are embedded into the vision encoder to produce interactive image representations \(\tilde{\mathbf{F}}_{v}\).
\(\tilde{h}^{f}\) is formed by concatenating \(h^{t}_{\textrm{eos}}\) and \(\tilde{h}^{v}_{\textrm{cls}}\).
\item \textbf{V2T:} Image representations \(\mathbf{F}_{v}\) are embedded into the text encoder to produce interactive text representations \(\tilde{\mathbf{F}}_{t}\).
\(\tilde{h}^{f}\) is formed by concatenating \(\tilde{h}^{t}_{\textrm{eos}}\) and \(h^{v}_{\textrm{cls}}\).
\item \textbf{Two-way (TW):} Both text and image representations \(\mathbf{F}_{t}\) and \(\mathbf{F}_{v}\) are embedded into each other's encoders, resulting in \(\tilde{\mathbf{F}}_{t}\) and \(\tilde{\mathbf{F}}_{v}\).
\(\tilde{h}^{f}\) is formed by concatenating \(\tilde{h}^{t}_{\textrm{eos}}\) and \(\tilde{h}^{v}_{\textrm{cls}}\).  
\end{itemize}
We will analyze the effectiveness of these three interaction modes in the experimental analysis.
\subsection{Training Strategy}
As shown in Figure~\ref{fig:framework} (left), to adapt InterCLIP for MEP, we introduce an efficient training strategy.
Using InterCLIP as the backbone to obtain fused features of the samples, we introduce a classification module and a projection module.

Given the fused features of a batch of samples \(\tilde{H}^{f} \in \mathbb{R}^{{N} \times (d_{t} + d_{v})}\), the classification module \(\mathcal{F}_{c}\) calculates the probabilities $\hat{y}$ of these samples being sarcastic or non-sarcastic:
\begin{equation}
\hat{y} = \textrm{softmax}(\mathcal{F}_{c}(\tilde{H}^{f})), \quad \hat{y} \in \mathbb{R}^{N \times 2},
\end{equation}
where \(N\) denotes the batch size.
We optimize \(\mathcal{F}_{c}\) using binary cross-entropy loss:
\begin{equation}
\mathcal{L}^{c} = - \frac{1}{N} \sum_{i=1}^N \left[ y_i \log(\hat{y}_{i,1}) + (1 - y_i) \log(1 - \hat{y}_{i,0}) \right],
\end{equation}
where \(y_{i}\) denotes the label of the \(i\)-th sample, with sarcastic labeled as 1 and non-sarcastic as 0, and \(\hat{y}_{i}\) denotes the prediction for the \(i\)-th sample.

The projection module \(\mathcal{F}_{p}\) maps \(\tilde{H}^{f}\) into a latent feature space:
\begin{equation}
\hat{H}^{f} = \textrm{norm}(\mathcal{F}_{p}(\tilde{H}^{f})), \quad \hat{H}^{f} \in \mathbb{R}^{{N} \times d_{f}},
\end{equation}
where \(\textrm{norm}(\cdot)\) denotes L2 normalization, and \(d_{f}\) represents the dimension of the projected features. 
In this space, the cosine distance between features of the same class is minimized, while the distance between features of different classes is maximized. 
We use a label-aware cosine similarity loss to optimize \(\mathcal{F}_{p}\):
\begin{equation}
\mathcal{L}^{p} = \textrm{mean}(\hat{H}^{f}_{P} \cdot \hat{H}^{f^{T}}_{N}) + \textrm{mean}(1 - \hat{H}^{f}_{P} \cdot \hat{H}^{f^{T}}_{P}) + \textrm{mean}(1 - \hat{H}^{f}_{N} \cdot \hat{H}^{f^{T}}_{N}),
\end{equation}
where \(\hat{H}^{f}_{P}\) and \(\hat{H}^{f}_{N}\) represent the projected features of positive and negative samples, respectively.

We fully train the modules \(\mathcal{F}_{c}\), \(\mathcal{F}_{p}\), the adapting projection layers (\(\mathcal{F}_{t}\) and \(\mathcal{F}_{v}\)), the gated projection layers (\(\mathcal{G}_{t}\) and \(\mathcal{G}_{v}\)), and the learnable gating parameters (\(\beta_{t}\) and \(\beta_{v}\)).
We use LoRA~\cite{edward2022lora} to fine-tune parts of the weight matrices \(\mathbf{W}\) in the self-attention modules of all encoders, specifically various combinations of \(W_{q}\), \(W_{k}\), \(W_{v}\), and \(W_{o}\).
We consider \(\mathbf{W}\) and the rank \(r\) of LoRA as hyperparameters for our study.
All learnable parts are optimized by minimizing the joint loss:
\begin{equation}
\mathcal{L} = \mathcal{L}^{c} + \mathcal{L}^{p}.
\end{equation}

\subsection{Memory-Enhanced Predictor}
As depicted in Figure~\ref{fig:framework} (right), we present the Memory-Enhanced Predictor (MEP) that builds upon the learned InterCLIP, along with the classification module and the projection module, leveraging the valuable historical knowledge of test samples to enhance multi-modal sarcasm detection.

\begin{wrapfigure}{r}{0.45\textwidth}
\tiny
\begin{minipage}{\linewidth}
\begin{algorithm}[H]
\caption{Memory-Enhanced Predictor}
\label{alg:mep-algorithm}
\begin{algorithmic}[1]
\STATE \textbf{Input}: Memory size $L$, Learned InterCLIP model, classification module $\mathcal{F}_{c}$ and projection module $\mathcal{F}_{p}$
\STATE \textbf{Output}: Final prediction $\hat{y}^{p}$
\STATE Initialize memory $\mathcal{M} \in \mathbf{0}^{2 \times L \times d_{f}}$
\STATE Initialize index $\mathcal{I} \in \mathbf{0}^{2}$
\STATE Initialize entropy records $\mathcal{C} \in \mathbf{0}^{2 \times L}$
\FOR{$i \gets 1$ \TO $N_{\textrm{test}}$}
    \STATE $\tilde{h}^{f}_{i} \gets \textrm{InterCLIP}(\mathcal{P}_{i})$
    \STATE $\hat{y}_{i} \gets \textrm{softmax}(\mathcal{F}_{c}(\tilde{h}^{f}_{i}))$
    \STATE $\ell_{\textrm{pse}_{i}} \gets \arg\max_{j}(\hat{y}_{i,j}), \; j\in\{0,1\}$
    \STATE $c_{i} \gets -\hat{y}_{i,0}\log{\hat{y}_{i,0}} - \hat{y}_{i,1}\log{\hat{y}_{i,1}}$
    \STATE $\hat{h}^{f}_{i} \gets \textrm{norm}(\mathcal{F}_{p}(\tilde{h}^{f}_{i}))$
    \IF{$\mathcal{I}[\ell_{\textrm{pse}_{i}}] < L$}
        \STATE $\mathcal{M}[\ell_{\textrm{pse}_{i}}][\mathcal{I}[\ell_{\textrm{pse}_{i}}]] \gets \hat{h}^{f}_{i}$
        \STATE $\mathcal{C}[\ell_{\textrm{pse}_{i}}][\mathcal{I}[\ell_{\textrm{pse}_{i}}]] \gets c_{i}$
        \STATE $\mathcal{I}[\ell_{\textrm{pse}_{i}}] \gets \mathcal{I}[\ell_{\textrm{pse}_{i}}] + 1$
    \ELSE
        \STATE $j \gets \textrm{GetMaxIdx}(\mathcal{C}[\ell_{\textrm{pse}_{i}}])$
        \IF{$c_{i} < \mathcal{C}[\ell_{\textrm{pse}_{i}}][j]$}
            \STATE $\mathcal{M}[\ell_{\textrm{pse}_{i}}][j] \gets \hat{h}^{f}_{i}$
            \STATE $\mathcal{C}[\ell_{\textrm{pse}_{i}}][j] \gets c_{i}$
        \ENDIF
    \ENDIF
    \STATE $\textrm{logits} \gets 
    \left[
        \sum_{k=0}^{\mathcal{I}[0]} (\hat{h}^{f}_{i}\mathcal{M}[0]^{T})_{k}, 
        \sum_{k=0}^{\mathcal{I}[1]} (\hat{h}^{f}_{i}\mathcal{M}[1]^{T})_{k}
    \right]$
    \STATE $\hat{y}^{p}_{i} \gets \textrm{softmax}(\textrm{logits})$
    \STATE \textbf{yield} $\hat{y}^{p}_{i}$
\ENDFOR
\end{algorithmic}
\end{algorithm}
\end{minipage}
\end{wrapfigure}

The detailed computational process of MEP is provided in Algorithm~\ref{alg:mep-algorithm}, where $N_{\textrm{test}}$ denotes the number of test samples.
MEP uses the trained InterCLIP to extract fused features of the samples. 
It utilizes the classification module $\mathcal{F}_{c}$ to assign a pseudo-label $\ell_{\textrm{pse}_{i}}$ to each sample $\mathcal{P}_{i}$ and the projection module $\mathcal{F}_{p}$ to obtain the sample's projected feature $\hat{h}^{f}_{i}$. 
To store valuable projected features of test samples as historical knowledge, MEP maintains a dynamic fixed-length dual-channel memory $\mathcal{M} \in \mathcal{R}^{2 \times L \times d_{f}}$, where $L$ is the memory length per channel. The first channel stores projected features of non-sarcastic samples, while the second channel stores those of sarcastic samples.
Based on the pseudo-label $\ell_{\textrm{pse}_{i}}$, the appropriate memory channel $\mathcal{M}[\ell_{\textrm{pse}_{i}}]$ is selected for updating.
If the selected channel has available space, the sample's projected features are added directly, and the prediction entropy is recorded.
If the memory is full, the prediction entropy of all samples in the memory is compared with that of the current sample.
Samples with the highest entropy are replaced, ensuring the retained samples have lower entropy.
Finally, the current sample's projected feature is combined with the historical features stored in both memory channels $\mathcal{M}$ using cosine similarity to yield the final prediction.

\begin{wraptable}{r}{0.44\textwidth}
\tiny
\centering
\caption{
Statistics of MMSD and MMSD2.0.
}
\label{tab:mmsd2.0-dataset}
\begin{tabularx}{\linewidth}{
l
l
l
l
}
\toprule
MMSD/MMSD2.0           & Sarcastic                & Non-sarcastic         & All                   \\
\midrule
Train                  & 8,642/9,576              & 11,174/10,240         & 19,816/19,816         \\
Validation             & 959/1,042                & 1,451/1,368           & 2,410/2,410           \\
Test                   & 959/1,037                & 1,450/1,372           & 2,409/2,409           \\
\bottomrule
\end{tabularx}
\end{wraptable}

\section{Experiments}
\label{sec:experimentsI}

\subsection{Experimental Settings}
\paragraph{Datasets and metrics.}
Following~\cite{qin2023mmsd2}, we evaluate performance on MMSD~\cite{Cai2019MultiModalSD} and MMSD2.0~\cite{qin2023mmsd2} using accuracy (Acc.), precision (P), recall (R), and F1-score (F1) as metrics.
We present the statistics of the two datasets in Table~\ref{tab:mmsd2.0-dataset}.

\paragraph{Baselines.}
We compare the effectiveness of the InterCLIP-MEP framework against several unimodal and multi-modal methods.
For text modality methods, we compare with TextCNN~\cite{kim-2014-convolutional}, Bi-LSTM~\cite{zhou-etal-2016-attention}, SMSD~\cite{xiong2019sarcasm}, and RoBERTa~\cite{liu2019roberta}.
For image modality methods, we compare with ResNet~\cite{he2015deep} and ViT~\cite{dosovitskiy2020image}.
We compare with state-of-the-art multi-modal methods, including HFM~\cite{Cai2019MultiModalSD}, Att-BERT~\cite{pan-etal-2020-modeling}, CMGCN~\cite{Liang2022MultiModalSD}, HKE~\cite{Liu2022TowardsMS}, DIP~\cite{wen2023dip}, DynRT~\cite{Tian2023DynamicRT}, Multi-view CLIP~\cite{qin2023mmsd2}, and G\textsuperscript{2}SAM~\cite{wei2024g2sam}, which employ various techniques such as hierarchical fusion, graph neural networks, and dynamic routing for multi-modal sarcasm detection.

\subsection{Implementation Details}
The model training and testing were conducted using PyTorch Lightning\footnote{\url{https://lightning.ai/}}. 
InterCLIP was constructed by leveraging the Transformers library~\cite{wolf-etal-2020-transformers}.
For the MMSD2.0 experiments, the initial weights for InterCLIP are based on \textsf{clip-vit-base-patch32}\footnote{\url{https://hf.co/openai/clip-vit-base-patch32}}.
For the MMSD experiments, we utilized the \textsf{roberta-ViT-B-32} model architecture provided by OpenCLIP\footnote{\url{https://github.com/mlfoundations/open_clip}}, with the pre-trained checkpoint \textsf{laion2b\_s12b\_b32k}\footnote{\url{https://hf.co/laion/CLIP-ViT-B-32-roberta-base-laion2B-s12B-b32k}}.
Custom scripts were developed to adapt its format to the Transformers library, ensuring compatibility with our framework.
The model parameters were optimized using AdamW~\cite{loshchilov2017decoupled}, with a learning rate set to 1e-4 for the LoRA fine-tuning modules and 5e-4 for other trainable modules. 
A cosine annealing scheduler with warmup was employed to dynamically adjust the learning rate, where the warmup steps constituted the first 20\% of the total optimization steps, and the minimum learning rate was set to 1\% of the initial rate.
For the modules $\mathcal{G}_{t/v}$, $\mathcal{F}_{t/v}$, $\mathcal{F}_{c}$, and $\mathcal{F}_{p}$, simple multi-layer perceptrons (MLPs) were utilized. 
The training processing was performed with a batch size of 64 for 3 epochs. 
All experiments were run on a machine equipped with an NVIDIA RTX 4090 GPU.

\begin{table}[ht]
\small
\centering
\begin{minipage}{0.54\textwidth}
\centering
\caption{
Summary of hyperparameters.
}\label{tab:hyperparameter-summary}
\begin{tabularx}{\linewidth}{lX}
\toprule
Parameter                                  & Description   \\
\midrule
$r$ 
& The rank of LoRA, determining the dimension of the low-rank update matrices.  \\

$\mathbf{W}$
& The weight matrices in the self-attention module fine-tuned using LoRA, specifically targeting combinations of $W_{\{q,k,v,o\}}$.\\

top-$n$
& The number of top self-attention layers conditioned during fine-tuning.\\

$d_{f}$
& The dimensionality of the latent space for the projected features. \\

$\mathbf{L}$
& The configurable range of memory sizes maintained by the Memory-Enhanced Predictor (MEP). \\
\bottomrule
\end{tabularx}
\end{minipage}%
\hfill
\begin{minipage}{0.43\textwidth}
\centering
\caption{
Hyperparameter settings.
}
\label{tab:hyperparameter-settings}
\begin{tabularx}{\linewidth}{lX}
\toprule
Parameter                                  & Value   \\
\midrule
\textit{For trainer}                      &\\
\cmidrule{1-1}
epoch                                     & 3    \\
batch\_size                                & 64   \\
lr                                        & 5e-4 \\
lora\_lr                                   & 1e-4 \\
warmup\_ratio                              & 0.2  \\
min\_lr\_rate                               & 0.01 \\
\midrule
\textit{For our model}                    &\\
\cmidrule{1-1}
$r$                                       & 8   \\
top-$n$                                   & 4    \\
$d_{f}$                                   & 1024   \\
$\mathbf{W}$                              & $W_{k},W_{v},W_{o}$   \\
$\mathbf{L}$                              & \{128, 256, 384, 512, 640, 768, 896, 1024, 1152, 1280\} \\
\bottomrule
\end{tabularx}
\end{minipage}
\end{table}

\subsection{Hyperparameter Details}
We summarize the hyperparameters involved in InterCLIP-MEP and their descriptions in Table~\ref{tab:hyperparameter-summary}.
The hyperparameter settings for obtaining the main results in the paper are summarized in Table~\ref{tab:hyperparameter-settings}.
For each experiment, we condition only the top four layers of the self-attention modules, with the projection dimension \(d_{f}\) set to 1024.
We set the LoRA rank \(r\) to 8, fine-tuning the self-attention weight matrices \(\mathbf{W}\), specifically \(W_{k}\), \(W_{v}\), and \(W_{o}\).
For the memory size \(L\), we select the optimal size from a fixed set of candidate values \(\mathbf{L}\).
For other baseline methods, we follow the optimal hyperparameter settings they reported.

\subsection{Main Results} \label{sec:main-results}
To begin our experimental analysis, we first evaluate the overall effectiveness of the proposed InterCLIP-MEP framework.
To further assess its robustness, we design a control variant, denoted as \texttt{w/o Inter}, where the original CLIP is used as the backbone in place of InterCLIP.
This variant is compared against three interaction modes of InterCLIP: \texttt{w/ V2T}, \texttt{w/ T2V}, and \texttt{w/ TW}.
The overall results and baseline comparisons are summarized in Table~\ref{tab:main-results}, while the effectiveness of different interaction modes is further analyzed in Table~\ref{tab:inter-results}.

\paragraph{Performance on MMSD2.0}
As shown in Table~\ref{tab:main-results}, our framework consistently surpasses or matches state-of-the-art baselines on MMSD2.0, confirming the effectiveness of the proposed training strategy and the Memory-Enhanced Predictor (MEP).
Table~\ref{tab:inter-results} further compares different interaction settings. Both \texttt{w/ V2T} and \texttt{w/ T2V} outperform the \texttt{w/o Inter} variant, indicating that InterCLIP more effectively captures cross-modal sarcasm cues than the original CLIP backbone.
Among them, \texttt{w/ T2V} yields the best results.
This is because projecting compact, low-dimensional text embeddings into the higher-dimensional vision space allows redundancy in visual features to accommodate textual semantics, preserving information more effectively.
In contrast, projecting rich, high-dimensional vision embeddings into the text space requires compression into a compact semantic space, which inevitably loses details and hampers alignment.
This asymmetry explains why \texttt{T2V} consistently outperforms \texttt{V2T}.
Meanwhile, the \texttt{w/ TW} configuration performs worse than either single-direction variant, as simultaneously embedding both modalities exacerbates the alignment asymmetry and further increases the learning burden.
In summary, InterCLIP with T2V interaction, together with our lightweight training strategy and MEP, achieves the most promising results on MMSD2.0, underscoring the robustness and adaptability of our framework in modeling subtle text–image interactions.

\begin{table}[t]
\centering
\scriptsize
\begin{threeparttable}
\caption{
Baseline comparison on MMSD2.0 and MMSD, where InterCLIP-MEP uses the best interaction mode for each dataset.
}
\label{tab:main-results}
\begin{tabularx}{\linewidth}{
X
|
>{\centering\arraybackslash}c
>{\centering\arraybackslash}c
>{\centering\arraybackslash}c
>{\centering\arraybackslash}c
|
>{\centering\arraybackslash}c
>{\centering\arraybackslash}c
>{\centering\arraybackslash}c
>{\centering\arraybackslash}c
}
\toprule
\multirow{2}{*}{Method}                                           & \multicolumn{4}{c|}{MMSD2.0}                                                                                                                                                                                         & \multicolumn{4}{c}{MMSD}                                                                                                                                                                              \\
\cmidrule(lr){2-5} \cmidrule(l){6-8}\cmidrule(r){9-9}
                                                                  & Acc. (\%)                                                & F1 (\%)                                           & P (\%)                                              & R (\%)                                          & Acc. (\%)                                                & F1 (\%)                                            & P (\%)                           & R (\%)                                             \\
\midrule
\multicolumn{9}{l}{\textit{Text}}                                                                                                                                                                                                                                                                                                                                                                                                                                                                 \\
\midrule
TextCNN~\cite{kim-2014-convolutional}                             & 71.61\textsuperscript{*}                                 & 69.52\textsuperscript{*}                           & 64.62\textsuperscript{*}                            & 75.22\textsuperscript{*}                       & 80.03\textsuperscript{*}                                 & 75.32\textsuperscript{*}                           & 74.29\textsuperscript{*}                     & 76.39\textsuperscript{*}                \\
Bi-LSTM~\cite{zhou-etal-2016-attention}                                      & 72.48\textsuperscript{*}                                 & 68.05\textsuperscript{*}                           & 68.02\textsuperscript{*}                            & 68.08\textsuperscript{*}                       & 81.90\textsuperscript{*}                                 & 77.53\textsuperscript{*}                           & 76.66\textsuperscript{*}                     & 78.42\textsuperscript{*}                \\
SMSD~\cite{xiong2019sarcasm}                                      & 73.56\textsuperscript{*}                                 & 69.97\textsuperscript{*}                           & 68.45\textsuperscript{*}                            & 71.55\textsuperscript{*}                       & 80.90\textsuperscript{*}                                 & 75.82\textsuperscript{*}                           & 76.46\textsuperscript{*}                     & 75.18\textsuperscript{*}                \\
RoBERTa~\cite{liu2019roberta}                                     & 79.66\textsuperscript{*}                                 & 76.21\textsuperscript{*}                           & 76.74\textsuperscript{*}                            & 75.70\textsuperscript{*}                       & \textbf{93.97}\textsuperscript{*}                        & \textbf{92.45}\textsuperscript{*}                  & \textbf{90.39}\textsuperscript{*}            & \textbf{94.59}\textsuperscript{*}       \\
\midrule
\multicolumn{9}{l}{\textit{Image}}                                                                                                                                                                                                                                                                                                                               \\
\midrule
ResNet~\cite{he2015deep}                                          & 65.50\textsuperscript{*}                                 & 57.58\textsuperscript{*}                           & 61.17\textsuperscript{*}                            & 54.39\textsuperscript{*}                       & 64.76\textsuperscript{*}                                 & 61.53\textsuperscript{*}                           & 54.41\textsuperscript{*}                     & 70.80\textsuperscript{*}                \\
ViT~\cite{dosovitskiy2020image}                                   & 72.02\textsuperscript{*}                                 & 69.72\textsuperscript{*}                           & 65.26\textsuperscript{*}                            & 74.83\textsuperscript{*}                       & 67.83\textsuperscript{*}                                 & 63.40\textsuperscript{*}                           & 57.93\textsuperscript{*}                     & 70.07\textsuperscript{*}                \\ 
\midrule
\multicolumn{9}{l}{\textit{Text-Image}}                                                                                                                                                                                                                                                                                                                                                                                                                                                            \\
\midrule
HFM~\cite{Cai2019MultiModalSD}                                    & 70.57\textsuperscript{*}                                 & 66.88\textsuperscript{*}                           & 64.84\textsuperscript{*}                            & 69.05\textsuperscript{*}                       & 83.44\textsuperscript{*}                                 & 80.18\textsuperscript{*}                           & 76.57\textsuperscript{*}                     & 84.15\textsuperscript{*}                 \\ 
Att-BERT~\cite{pan-etal-2020-modeling}                            & 80.03\textsuperscript{*}                                 & 77.04\textsuperscript{*}                           & 76.28\textsuperscript{*}                            & 77.82\textsuperscript{*}                       & 86.05\textsuperscript{*}                                 & 82.92\textsuperscript{*}                           & 80.87\textsuperscript{*}                     & 85.08\textsuperscript{*}                 \\ 
CMGCN~\cite{Liang2022MultiModalSD}                                & 79.83\textsuperscript{*}                                 & 76.90\textsuperscript{*}                           & 75.82\textsuperscript{*}                            & 78.01\textsuperscript{*}                       & 86.54\textsuperscript{*}                                 & 84.09\textsuperscript{*}                           & -\textsuperscript{*}                         & -\textsuperscript{*}                     \\ 
HKE~\cite{Liu2022TowardsMS}                                       & 76.50\textsuperscript{*}                                 & 72.25\textsuperscript{*}                           & 73.48\textsuperscript{*}                            & 71.07\textsuperscript{*}                       & 87.36\textsuperscript{*}                                 & 72.25\textsuperscript{*}                           & 81.84\textsuperscript{*}                     & 86.48\textsuperscript{*}                 \\ 
DIP~\cite{wen2023dip}                                             & 80.59\textsuperscript{\dag}                              & 78.23\textsuperscript{\dag}                        & 75.52\textsuperscript{\dag}                         & 81.14\textsuperscript{\dag}                    & 89.59\textsuperscript{\dag}                              & 87.17\textsuperscript{\dag}                        & 87.76\textsuperscript{\dag}                  & 86.58\textsuperscript{\dag}              \\ 
DynRT~\cite{Tian2023DynamicRT}                                    & 70.37\textsuperscript{\dag}                             & 68.55\textsuperscript{\dag}                        & 63.02\textsuperscript{\dag}                          & 75.15\textsuperscript{\dag}                    & \underline{93.59}\textsuperscript{\dag}                  & \underline{91.93}\textsuperscript{\dag}            & \underline{90.30}\textsuperscript{\dag}      & \underline{93.62}\textsuperscript{\dag}  \\ 
Multi-view CLIP~\cite{qin2023mmsd2}                               & \underline{85.64}\textsuperscript{*}                    & \underline{84.10}\textsuperscript{*}               & \underline{80.33}\textsuperscript{*}                 & \underline{88.24}\textsuperscript{*}           & 88.33\textsuperscript{*}                                 & 85.55\textsuperscript{*}                           & 82.66\textsuperscript{*}                     & 88.65\textsuperscript{*}                 \\
G\textsuperscript{2}SAM~\cite{wei2024g2sam}                       & 79.43\textsuperscript{\dag}                            & 78.07\textsuperscript{\dag}                         & 72.04\textsuperscript{\dag}                          & 85.20\textsuperscript{\dag}                    & 90.48\textsuperscript{\dag}                              & 88.48\textsuperscript{\dag}                        & 87.95\textsuperscript{\dag}                  & 89.02\textsuperscript{\dag}              \\ 
\midrule
InterCLIP-MEP (Ours)   & \textbf{86.72}                                        & \textbf{85.61}                                       & 80.20                                                & \textbf{91.80}                                  & \textbf{93.94}                                          & \textbf{92.54}                                      & \textbf{90.77}                              & \textbf{94.37}                          \\
\bottomrule
\end{tabularx}
\begin{tablenotes}
\footnotesize
\raggedright
\item[\textasteriskcentered] We use $*$ to indicate that the results are taken from~\cite{qin2023mmsd2}.
\- indicates that results are not reported.
$\dagger$ indicates our reproduced results.
\underline{Underlined} values represent the best multi-modal baseline for comparison.
\textbf{Bold} values indicate those that surpass the underlined baseline.
The InterCLIP-MEP results use the T2V interaction on MMSD2.0 and the w/o Inter setting on MMSD. 
These choices follow our analysis (Section~\ref{sec:main-results}, Table~\ref{tab:inter-results}), which shows T2V is most effective on MMSD2.0, 
while MMSD’s spurious textual cues make w/o Inter more suitable.
\end{tablenotes}
\end{threeparttable}
\end{table}

\begin{table}[t]
\centering
\scriptsize
\begin{threeparttable}
\caption{
Performance comparison of InterCLIP-MEP under different interaction settings.
}
\label{tab:inter-results}
\begin{tabularx}{\linewidth}{
X
|
>{\centering\arraybackslash}c
>{\centering\arraybackslash}c
>{\centering\arraybackslash}c
>{\centering\arraybackslash}c
|
>{\centering\arraybackslash}c
>{\centering\arraybackslash}c
>{\centering\arraybackslash}c
>{\centering\arraybackslash}c
}
\toprule
\multirow{2}{*}{Method}                                           & \multicolumn{4}{c|}{MMSD2.0}                                                                                                                                                                                         & \multicolumn{4}{c}{MMSD}                                                                                                                                                                              \\
\cmidrule(lr){2-5} \cmidrule(l){6-8}\cmidrule(r){9-9}
                                                                  & Acc. (\%)                                                & F1 (\%)                                           & P (\%)                                              & R (\%)                                          & Acc. (\%)                                                & F1 (\%)                                            & P (\%)                           & R (\%)                                             \\
\midrule
\multicolumn{9}{l}{\textit{InterCLIP-MEP}}                                                                                                                                                                                                                                                                                                                                                                                                                                                  \\
\midrule
w/o Inter ($L_{2}=1024,L_{1}=1280$)                               & \textbf{86.05}                                        & \textbf{84.81}                                       & 79.83                                                & \textbf{90.45}                                   & 88.75                                                    & 86.31                                              & 83.73                                       & 89.05                                    \\
w/ TW ($L_{2}=128,L_{1}=1152$)                                    & 85.51                                                 & \textbf{84.26}                                       & 79.15                                                & \textbf{90.07}                                   & 88.54                                                    & 86.32                                              & 82.25                                       & 90.82                                    \\
w/ V2T ($L_{2}=640,L_{1}=1024$)                                   & \textbf{86.26}                                        & \textbf{85.00}                                       & 80.17                                                & \textbf{90.45}                                   & 88.92                                                    & 86.66                                              & 83.21                                       & 90.41                                    \\
w/ T2V ($L_{2}=1024,L_{1}=1152$)                                  & \textbf{86.72}                                        & \textbf{85.61}                                       & 80.20                                                & \textbf{91.80}                                   & 88.83                                                    & 86.37                                              & 84.02                                       & 88.84                                    \\
\midrule
\multicolumn{9}{l}{\textit{InterCLIP-MEP w/ RoBERTa}}                                                                                                                                                                                                                                                                                                                                                                                                                                       \\
\midrule
w/o Inter ($L_{2}=640,L_{1}=128$)                               & 77.21                                                 & 75.55                                                 & 70.20                                               & 81.77                                             & \textbf{93.94}                                          & \textbf{92.54}                                      & \textbf{90.77}                              & \textbf{94.37}                          \\
w/ TW ($L_{2}=896,L_{1}=256$)                                    & 81.98                                                 & 80.78                                                 & 74.69                                               & 87.95                                             & \textbf{93.73}                                          & \textbf{92.28}                                      & \textbf{90.48}                              & \textbf{94.16}                          \\
w/ V2T ($L_{2}=640,L_{1}=512$)                                   & 76.96                                                 & 75.26                                                 & 69.98                                               & 81.39                                             & \textbf{93.94}                                          & \textbf{92.54}                                      & \textbf{90.69}                              & \textbf{94.47}                          \\
w/ T2V ($L_{2}=1024,L_{1}=384$)                                  & 82.81                                                 & 81.55                                                 & 75.81                                               & 88.24                                             & \textbf{93.73}                                          & \textbf{92.28}                                      & \textbf{90.56}                              & \textbf{94.06}                          \\
\bottomrule
\end{tabularx}
\begin{tablenotes}
\footnotesize
\raggedright
\item[\textasteriskcentered] \textbf{Bold} values indicate those that surpass the underlined baseline.
$L_2$ for MMSD2.0 and $L_1$ for MMSD denote the optimal MEP memory sizes.
\end{tablenotes}
\end{threeparttable}
\end{table}

\paragraph{Performance on MMSD}

\begin{wraptable}{r}{0.35\textwidth}
\vspace{-2em}
\centering
\small
\caption{
Performance comparison on the MMSD dataset in terms of Macro-average F1 score, following the standard evaluation protocol used in recent literature~\cite{zhuang2025multi}.
Results marked with * are reported from~\cite{zhuang2025multi}.
}
\label{tab:mmsd-macro-results}
\begin{tabularx}{\linewidth}{Xc}
\toprule
Method                                                    & Macro-F1~(\%) \\
\midrule
DIP~\cite{wen2023dip}*                                    & 89.01 \\
G\textsuperscript{2}SAM~\cite{wei2024g2sam}*              & 89.65 \\
KFGC-Net~\cite{zhuang2025multi}*                          & 91.02 \\
InterCLIP-MEP                                             & 93.72  \\
\bottomrule
\end{tabularx}
\end{wraptable}

As shown in Table~\ref{tab:main-results}, the RoBERTa-based text baseline significantly outperforms other methods on MMSD due to spurious textual cues, which allow models to achieve high accuracy using text features alone~\cite{qin2023mmsd2}.
Consequently, multi-modal models such as DynRT~\cite{Tian2023DynamicRT}, G\textsuperscript{2}SAM~\cite{wei2024g2sam}, and DIP~\cite{wen2023dip}, which rely on RoBERTa or BERT for text encoding, also obtain strong results on MMSD but exhibit a substantial performance drop on MMSD2.0. To further investigate this effect, we introduce an additional variant, \textit{InterCLIP-MEP w/ RoBERTa}, by replacing the original text encoder with RoBERTa (Table~\ref{tab:inter-results}).
Although this variant achieves state-of-the-art performance on MMSD, it disrupts InterCLIP’s modality alignment and thus leads to a noticeable decline on MMSD2.0.
These findings suggest that MMSD’s text data contains spurious cues that encourage over-reliance on textual signals, whereas MMSD2.0, being cleaned, requires stronger multi-modal reasoning.
We also observe that the \texttt{w/ V2T} variant performs best in both \textit{InterCLIP-MEP} and \textit{InterCLIP-MEP w/ RoBERTa}, highlighting the model’s tendency to over-depend on textual features.
Following~\cite{zhuang2025multi}, we also report the Macro-F1 score on the MMSD dataset to ensure consistency with community evaluation standards. As shown in Table~\ref{tab:mmsd-macro-results}, InterCLIP-MEP achieves the highest Macro-F1 of 93.72\%, outperforming DIP~\cite{wen2023dip}, G$^2$SAM~\cite{wei2024g2sam}, and KFGC-Net~\cite{zhuang2025multi}.
\begin{table}[t]
\scriptsize
\centering
\begin{threeparttable}
\caption{
Efficiency comparison of different methods.
}
\label{tab:efficiency-comparison}
\begin{tabularx}{0.89\textwidth}{
l
l
l
l
l
l
}
\toprule
Method & \makecell{Accuracy \\ (\%)} & \makecell{Trainable Parameters \\ (M)} & \makecell{Fitting Time / Epoch \\ (s)} & \makecell{Inference Time \\ (s)} & \makecell{GPU Memory Peak \\ (GB)} \\
\midrule
Multi-view CLIP & \percentbarri{85.64}{4.6}{mygreen}{3} & \percentbarri{165}{8}{myblue}{5} & \percentbarri{488}{9}{myblue}{5} & \percentbarri{51}{7}{myblue}{3.5} & \percentbarri{15.59}{7.6}{myblue}{5} \\
DIP & \percentbarri{80.59}{4.3}{mygreen}{3} & \percentbarri{196}{9}{myblue}{5} & \percentbarri{\textcolor{red}{OOM}}{9}{myblue}{5.5} & \percentbarri{\textcolor{red}{OOM}}{7}{myblue}{4.3} & \percentbarri{\textcolor{red}{OOM}}{8}{myblue}{5.2} \\
G2SAM & \percentbarri{79.43}{4.2}{mygreen}{3} & \percentbarri{116}{7}{myblue}{5} & \percentbarri{90}{4}{myblue}{4.8} & \percentbarri{13}{2}{myblue}{3.5} & \percentbarri{18.32}{8}{myblue}{5} \\
DynRT & \percentbarri{70.37}{3}{mygreen}{3} & \percentbarri{25}{4}{myblue}{4.8} & \percentbarri{370}{7}{myblue}{5} & \percentbarri{26}{4}{myblue}{3.5} & \percentbarri{8.03}{3.5}{myblue}{4.9} \\
\rowcolor{gray!20}
InterCLIP-MEP & \percentbarri{86.72}{5}{mygreen}{3} & \percentbarri{8}{1.5}{myblue}{4.6} & \percentbarri{55}{2.6}{myblue}{4.8} & \percentbarri{6}{1}{myblue}{3.4} & \percentbarri{6.14}{3}{myblue}{4.9} \\
\bottomrule
\end{tabularx}
\begin{tablenotes}
\footnotesize
\raggedright
\item[\textasteriskcentered] To demonstrate the efficiency of InterCLIP-MEP, we selected several recent baselines for comparison.
The analysis was conducted using the MMSD2.0 dataset on a single NVIDIA RTX 4090 GPU with a batch size of 128.
In the table, \texttt{Fitting Time / Epoch} indicates the time required for each epoch during training and validation and \texttt{OOM} indicates Out of Memory, referring to GPU memory overflow.
\end{tablenotes}
\end{threeparttable}
\end{table}

\paragraph{Efficiency Comparison.}
Our training strategy demonstrates both remarkable effectiveness and outstanding efficiency.
To validate this, we conducted a comprehensive comparative analysis against leading state-of-the-art methods, as detailed in Table~\ref{tab:efficiency-comparison}.
For instance, the Multi-view CLIP method~\cite{qin2023mmsd2} employs a multi-layer Transformer encoder for feature fusion, which, while effective, introduces a significant number of trainable parameters.
This results in slower training and inference speeds and greater memory consumption.
Similarly, the DIP method~\cite{wen2023dip} caches historical samples during training, which hinders its ability to support large-batch training under limited resource conditions.
In contrast, our method operates with a batch size of 128 while utilizing a significantly smaller number of trainable parameters, which translates to notably faster training and validation cycles.
Furthermore, by incorporating minimal parameter modifications to adapt CLIP and utilizing simple yet effective linear layers for representation fusion, our approach achieves superior inference speeds and drastically reduced memory consumption.
These results highlight the practicality of our framework, establishing it as a benchmark for both computational efficiency and performance in multi-modal sarcasm detection.

\begin{table}[t]
\small
\centering
\begin{threeparttable}
\caption{
Ablation study of InterCLIP-MEP.
}
\label{tab:ablation-study}
\begin{tabularx}{0.89\textwidth}{
l
|
>{\centering\arraybackslash}c
>{\centering\arraybackslash}X
|
>{\centering\arraybackslash}c
>{\centering\arraybackslash}X
|
>{\centering\arraybackslash}c
>{\centering\arraybackslash}X
|
>{\centering\arraybackslash}c
>{\centering\arraybackslash}X
}
\toprule
                                        & \multicolumn{2}{c|}{\textit{w/o Inter}}                                       & \multicolumn{2}{c|}{\textit{w/ TW}}                                & \multicolumn{2}{c|}{\textit{w/ V2T}}                               & \multicolumn{2}{c}{\textit{w/ T2V}}                         \\
\midrule
Variant                                 & Acc. (\%)                              & F1 (\%)                             & Acc. (\%)                             & F1 (\%)                   & Acc. (\%)                             & F1 (\%)                   & Acc. (\%)                             & F1 (\%)             \\
\midrule
\textsc{Baseline}                       & \textbf{86.05}                         & \textbf{84.81}                      & \textbf{85.51}                        & \textbf{84.26}            & \textbf{86.26}                        & \textbf{85.00}            & \textbf{86.72}                        & \textbf{85.61}      \\
w/o Proj                                & 85.76                                  & 84.43                               & 85.43                                 & 84.05                     & 85.68                                 & 84.22                     & 86.22                                 & 84.51               \\
w/o MEP                                 & 85.39	                                 & 83.99                               & 85.22	                               & 83.79                     & 86.26	                               & 84.78                     & 86.26	                               & 84.82               \\
w/o LoRA                                & 82.44		                             & 77.73                               & 76.42		                           & 74.37                     & 73.31		                           & 72.22                     & 75.13		                           & 71.79               \\
\bottomrule
\end{tabularx}
\begin{tablenotes}
\footnotesize
\raggedright
\item[\textasteriskcentered] \textsc{Baseline} denotes the results without ablation.
\end{tablenotes}
\end{threeparttable}
\end{table}

\subsection{Analysis of InterCLIP-MEP}
To robustly validate the effectiveness of InterCLIP-MEP, we conduct comprehensive ablation studies and case studies on the more reliable MMSD2.0 benchmark, offering deeper insights into its design and performance.
In addition, we include visualization analyses to provide an intuitive understanding of how the framework processes multi-modal sarcasm cues.

\paragraph{Ablation study.}
We remove the projection module \(\mathcal{F}_p\) and train only the classification module \(\mathcal{F}_c\) for prediction, denoted as \texttt{w/o Proj}.
To test the necessity of using LoRA~\cite{edward2022lora} for fine-tuning, we keep the rest of InterCLIP-MEP unchanged and freeze all self-attention weights of InterCLIP, denoted as \texttt{w/o LoRA}, always selecting the optimal memory size \(L\) for the MEP during inference.
To evaluate the effectiveness of the MEP, we train both \(\mathcal{F}_p\) and \(\mathcal{F}_c\) but use only \(\mathcal{F}_c\) during inference, denoted as \texttt{w/o MEP}.
Table~\ref{tab:ablation-study} reports all results. 
All variants show performance declines compared to the baseline, demonstrating the importance of each module in InterCLIP-MEP. 
For \texttt{w/ TW} and \texttt{w/o Inter}, the \texttt{w/o MEP} variant performed worse than the \texttt{w/o Proj} variant.
However, for \texttt{w/ T2V} and \texttt{w/ V2T}, the \texttt{w/o MEP} variant performs better than the \texttt{w/o Proj} variant. 
This suggests that backbones with strong image-text interaction capabilities benefit from training the classification module $\mathcal{F}_c$ along with the projection module $\mathcal{F}_p$, even without using MEP during inference.
We also find that not using LoRA to fine-tune the self-attention modules results in significant performance loss, indicating that the original CLIP's vanilla space is not suitable for the sarcasm detection task.

\begin{figure}[t]
\centering
\begin{minipage}[c]{0.48\textwidth}
\centering
\includegraphics[width=\linewidth]{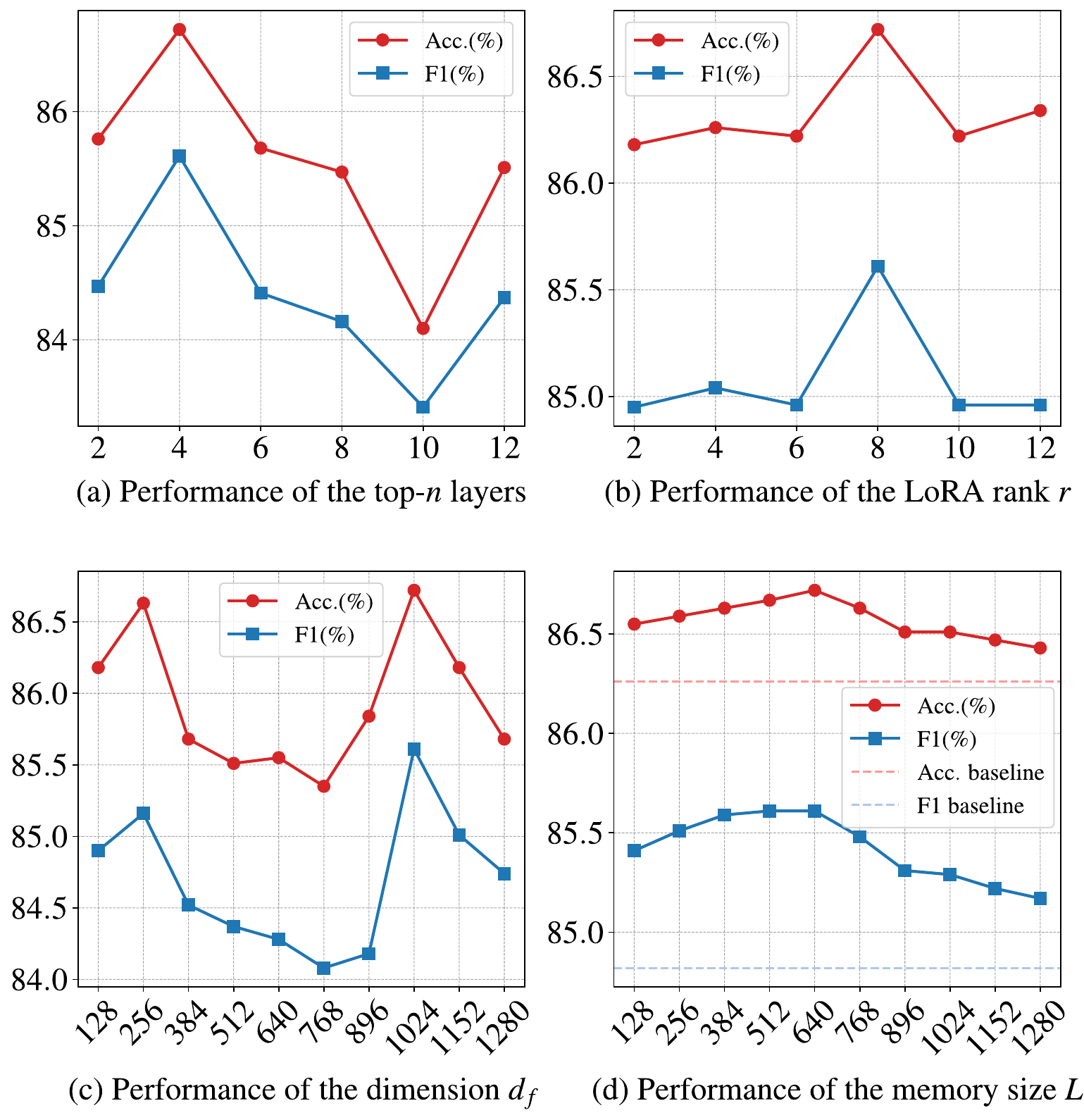}
\caption{
Hyperparameter study curves for \texttt{w/ T2V}.
Panel (d) compares results with those from using only the classification module $\mathcal{F}_{c}$ for prediction.
}
\label{fig:n-d_f-r-L}
\end{minipage}\hfill
\begin{minipage}[c]{0.48\textwidth}
\centering
\includegraphics[width=\linewidth]{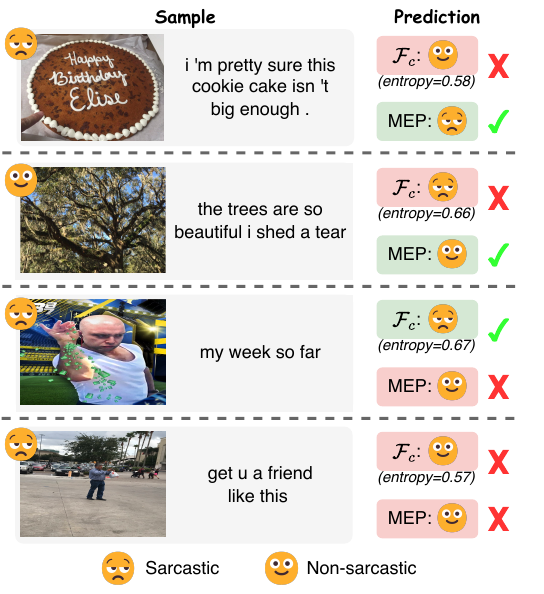}
\caption{
Case study of InterCLIP-MEP.
In the figure, the emojis in the \textit{Sample} column denote the ground-truth labels from the dataset. \textit{MEP} represents the labels predicted by the memory-enhanced predictor, and $\mathcal{F}_{c}$ represents the labels predicted by the classification module.
}
\label{fig:case-study}
\end{minipage}
\end{figure}

\paragraph{Hyperparameter study.}
We further investigate the method using Interactive-CLIP with T2V interaction as the backbone.
Keeping the other hyperparameters constant, we condition different top-$n$ layers of the self-attention modules.
We also study the impact of different projection dimensions $d_{f}$, different LoRA ranks $r$, and different memory sizes $L$ on the \texttt{w/ T2V} method.
We present all results in Figure~\ref{fig:n-d_f-r-L}.
Figure~\ref{fig:n-d_f-r-L}(a) shows that conditioning the top four self-attention layers yields the best results. 
From Figure~\ref{fig:n-d_f-r-L}(b), a rank of 8 is optimal. 
Figure~\ref{fig:n-d_f-r-L}(c) indicates the projection dimension is best at 256 or 1024. 
Figure~\ref{fig:n-d_f-r-L}(d) reveals that a memory size of 640 in MEP outperforms the others, confirming the value of historical sample knowledge.

\begin{figure}[t]
\centering
\includegraphics[width=0.97\textwidth]{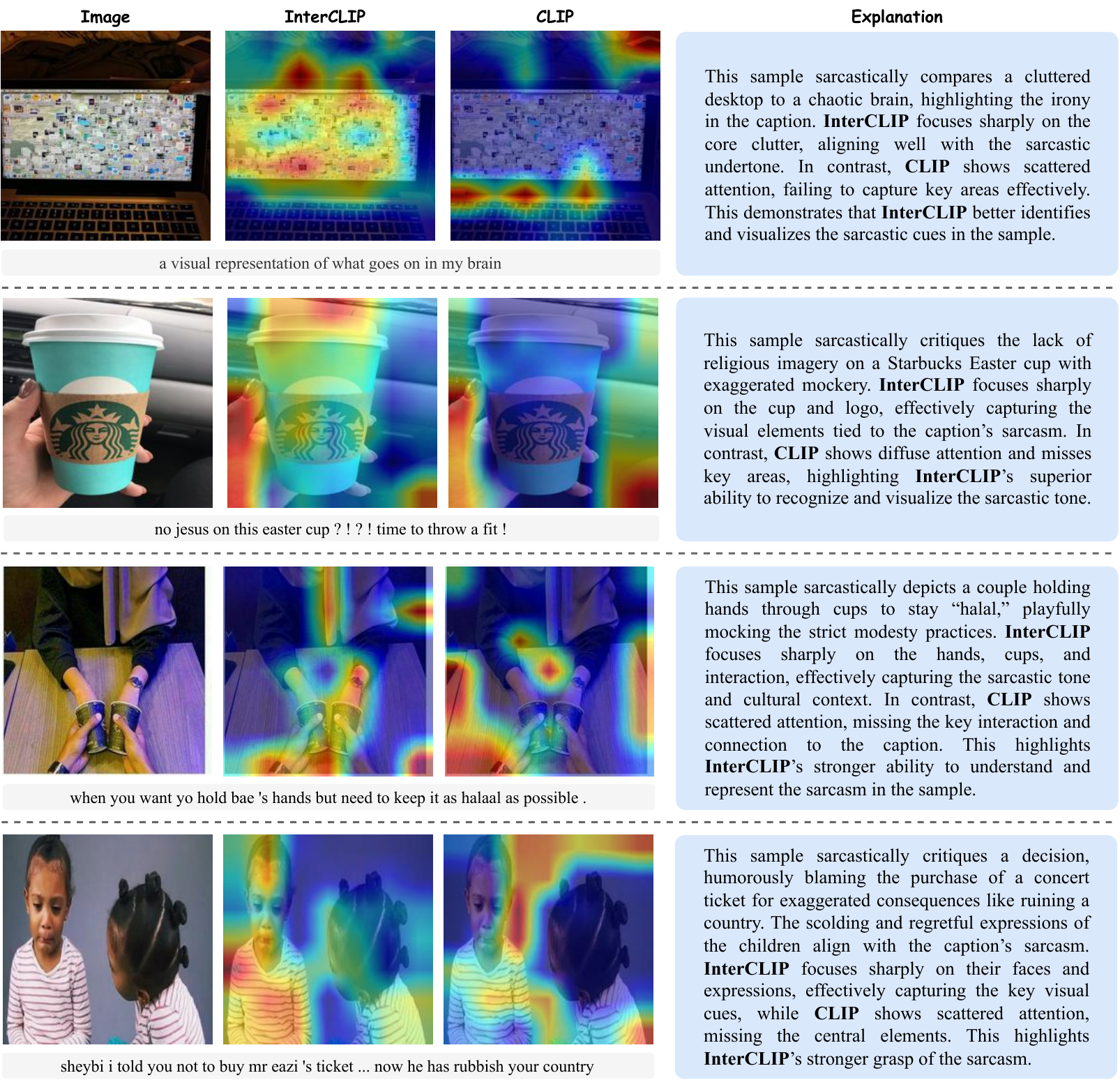}
\caption{
Visual examples showcasing InterCLIP's improved focus on sarcasm-related visual cues compared to CLIP.
In the figure, the first column shows the original input images, the second and third columns present the GradCAM visualizations from InterCLIP and the original CLIP respectively, and the last column provides human-readable explanations comparing their behaviors.
}
\label{fig:visualization}
\end{figure}

\paragraph{Case study.}
As shown in Figure~\ref{fig:case-study}, we provide illustrative examples to examine both the strengths and the limitations of InterCLIP-MEP.
In the first two cases, direct predictions from the classification module $\mathcal{F}_{c}$ yield uncertain and incorrect outcomes with relatively high entropy, whereas the Memory-Enhanced Predictor (MEP) leverages historical sample knowledge to successfully correct the results.
These examples highlight the robustness and effectiveness of our design.

At the same time, we also include two challenging failure cases.
In the third example, $\mathcal{F}_{c}$ makes a correct prediction, while MEP mistakenly overrides it.
This situation arises when the memory stores similar but misleading representations, which can bias the final decision.
In the fourth example, both $\mathcal{F}_{c}$ and MEP fail, reflecting the intrinsic difficulty of multi-modal sarcasm detection when subtle ironic cues are either extremely implicit or context-dependent beyond the available modalities.
These observations naturally raise two questions: how does the memory mechanism handle potential error propagation, and how robust is it under distribution shifts?
To answer these, we provide dedicated analyses in Section~\ref{sec:err-prop-exp} and Section~\ref{sec:dist-shift-exp}, where we examine these challenges in a systematic manner.

With these results in mind, it is important to note that the identified failures do not undermine the overall effectiveness of InterCLIP-MEP, but rather point to promising directions for refinement.
For instance, incorporating context-aware memory update strategies or external commonsense knowledge may further reduce such errors.
Overall, the inclusion of both successful and unsuccessful cases provides a more balanced perspective, demonstrating that while our method substantially improves robustness, multi-modal sarcasm detection remains a highly challenging task.

\paragraph{Visualization.}
To further validate that InterCLIP is capable of more effectively capturing  the interactive information between text and images compared to original CLIP, thereby aiding in the detection of sarcasm cues, we use GradCAM~\cite{Selvaraju2017GradCAM} to visualize the areas of focus during the inference process of the visual model in Figure~\ref{fig:visualization}.
These visualizations highlight InterCLIP’s improved ability to capture sarcasm-related cues by focusing on relevant areas in the images.

\subsection{Empirical Study of Self-attention Fine-tuning and Interaction Modes}
Keeping the other hyperparameters constant as shown in Table~\ref{tab:hyperparameter-settings}, we fine-tune all possible weight matrices $\mathbf{W}$ and employ different interaction modes of InterCLIP as the backbone.
We consistently select the optimal memory size from $\mathbf{L}$ for MEP.
We calculate the average metrics for different methods and different weight matrices.
The results are presented in Table~\ref{tab:lora-module-exp-w} and Table~\ref{tab:lora-module-exp-met}.
We observe that fine-tuning the weight matrices $W_{k}, W_{v}, W_{o}$ and using the \textsf{T2V} interaction mode of InterCLIP are the best choices for InterCLIP-MEP.

\begin{table}[t]
\centering
\begin{minipage}{0.45\textwidth}
\centering
\caption{
Average results of fine-tuning different weight matrices $\mathbf{W}$ across four baseline methods.
}
\label{tab:lora-module-exp-w}
\begin{tabularx}{\columnwidth}{Xcc}
\toprule
$\mathbf{W}$                              & Mean Acc. (\%)                 & Mean F1 (\%)              \\
\midrule
$W_{q}$                                   & 85.14 	                       & 83.99                     \\
$W_{k}$                                   & 85.09 	                       & 84.00                     \\
$W_{v}$                                   & 85.34 	                       & 84.24                     \\
$W_{o}$                                   & 85.39 	                       & 84.23                     \\
$W_{q},W_{k}$                             & 85.36 	                       & 84.16                     \\
$W_{q},W_{v}$                             & 85.63 	                       & 84.40                     \\
$W_{q},W_{o}$                             & 85.73 	                       & 84.49                     \\
$W_{k},W_{o}$                             & 85.67 	                       & 84.43                     \\
$W_{v},W_{o}$                             & 85.73 	                       & 84.54                     \\
$W_{k},W_{v}$                             & 85.70 	                       & 84.51                     \\
$W_{q},W_{k},W_{o}$                       & 85.92 	                       & 84.63                     \\
$W_{q},W_{v},W_{o}$                       & 85.87 	                       & 84.60                     \\
$W_{q},W_{k},W_{v}$                       & 86.05 	                       & 84.75                     \\
$W_{k},W_{v},W_{o}$                       & \textbf{86.14} 	               & \textbf{84.92}                     \\
$W_{q},W_{k},W_{v},W_{o}$                 & 85.82 	                       & 84.55                     \\
\bottomrule
\end{tabularx}
\end{minipage}%
\hfill
\begin{minipage}{0.47\textwidth}
\centering
\caption{
Average results of four baseline methods for fine-tuning different weight matrices $\mathbf{W}$.
}
\label{tab:lora-module-exp-met}
\begin{tabularx}{\columnwidth}{Xcc}
\toprule
Method                      & Mean Acc. (\%)                   & Mean F1 (\%)               \\
\midrule
w/o Inter                   & 85.49 	                       & 84.32                      \\
w/ TW                       & 85.66 	                       & 84.44                      \\
w/ V2T                      & 85.62 	                       & 84.41                      \\
w/ T2V                      & \textbf{85.78} 	               & \textbf{84.55}                      \\
\bottomrule
\end{tabularx}

\vspace{5em}

\centering
\caption{Statistics of DocMSU.}
\label{tab:docmsu-statistics}
\begin{tabularx}{\linewidth}{Xccc}
\toprule
& Sarcastic & Non-sarcastic & All \\
\midrule
Train      & 4,014 & 46,265 & 50,279 \\
Validation & 1,125 & 13,097 & 14,222 \\
Test       & 555   & 6,772  & 7,327 \\
\bottomrule
\end{tabularx}
\end{minipage}
\end{table}

\section{Extended Experiments}\label{sec:experimentsII}
To further verify the performance and robustness of our framework, we extended our evaluation in three complementary directions.
First, we conducted experiments on a document-level benchmark to examine its capability in handling long-text sarcasm detection.
Second, we performed low-resource experiments to assess generalization and adaptability under data scarcity.
Third, we systematically investigated the impact of our memory mechanism under error propagation and distribution shift conditions, thereby providing deeper insights into its stability and reliability.

\subsection{Document-level Benchmark}
\paragraph{DocMSU Benchmark}
DocMSU~\cite{du2024docmsu} is a recently introduced multi-modal sarcasm benchmark designed specifically for long-text analysis.
It facilitates the evaluation of multi-modal sarcasm comprehension as well as detection tasks.
In this study, we concentrate on the multi-modal sarcasm detection task.
The benchmark statistics can be found in Table~\ref{tab:docmsu-statistics}.

\paragraph{Baselines}
We follow the evaluation protocol of~\cite{du2024docmsu}.
We compare against unimodal baselines: BERT-base (text-only)~\cite{devlin2019bert} and Swin Transformer (image-only)~\cite{liu2021swin}.
For multi-modal approaches, we include CLIP~\cite{radford2021learning}, Vision-and-Language Transformer (ViLT)~\cite{kim2021vilt}, and CMGCN~\cite{Liang2022MultiModalSD}.
The method proposed by~\cite{du2024docmsu} is taken as the state-of-the-art baseline.

\paragraph{Results}

\begin{wraptable}{r}{0.52\textwidth}
\centering
\small
\begin{threeparttable}
\caption{
Results on DocMSU.
}
\label{tab:docmsu-results}
\begin{tabularx}{\linewidth}{Xcccc}
\toprule
Method                                                    & Acc.                     & F1                    & P                   & R                \\
\midrule
BERT-base*                                                & 87.12                    & 86.51                 & 77.61                & 70.37                 \\
Swin-Transformer*                                         & 74.83                    & 61.51                 & 67.57                & 56.45                 \\
CMGCN*                                                    & 88.12                    & 75.23                 & 78.11                & 72.55                 \\
CLIP*                                                     & 96.19                    & 77.62                 & 78.99                & 76.30                 \\
ViLT*                                                     & 93.15                    & 41.44                 & 69.03                & 29.61                 \\
DocMSU Baseline*      & \underline{97.83}        & \underline{87.25}     & \underline{81.20}    & \underline{94.27}     \\
\midrule
\multicolumn{5}{l}{\textit{InterCLIP-MEP (Ours)}}  \\
\midrule
w/o Inter                                                & \textbf{97.84}           & \textbf{87.48}        & 78.08                & \textbf{99.45}        \\
w/ TW                                                    & 97.79                    & 87.24                 & 77.48                & \textbf{99.81}        \\
w/ V2T                                                   & 97.83                    & \textbf{87.45}        & 77.81                & \textbf{99.82}        \\
w/ T2V                                                   & 97.67                    & 86.65                 & 76.45                & \textbf{99.99}        \\
\bottomrule
\end{tabularx}
\begin{tablenotes}
\footnotesize
\item[\textasteriskcentered] \underline{Underline} results denote the compared SOTA baseline, \textbf{boldface} highlights results that surpass the baseline, and * indicates results sourced from~\cite{du2024docmsu}.
\end{tablenotes}
\end{threeparttable}
\end{wraptable}

Our InterCLIP-MEP framework demonstrates strong performance across various configurations, as shown in Table~\ref{tab:docmsu-results}.
In particular, the \texttt{w/o Inter} and \texttt{w/ V2T} variants consistently achieve higher F1 scores compared to the baselines, thereby showcasing their robustness in handling multi-modal sarcasm detection tasks.
Notably, all variants achieve nearly perfect recall, highlighting their outstanding capability in accurately identifying sarcasm across diverse datasets. 
The \texttt{w/o Inter} variant also achieves the highest accuracy, further demonstrating its effectiveness and precision.
Overall, the comprehensive results in Table~\ref{tab:docmsu-results} affirm the unparalleled effectiveness of InterCLIP's modality interaction mechanism and our proposed memory-enhanced predictor (MEP), particularly in surpassing existing state-of-the-art methods in both accuracy and F1 score, thus setting a new benchmark in the field.

\subsection{Low-resource Setting}
\begin{figure}[t]
\centering
\begin{minipage}[c]{0.4\textwidth}
\centering
\includegraphics[width=0.75\linewidth]{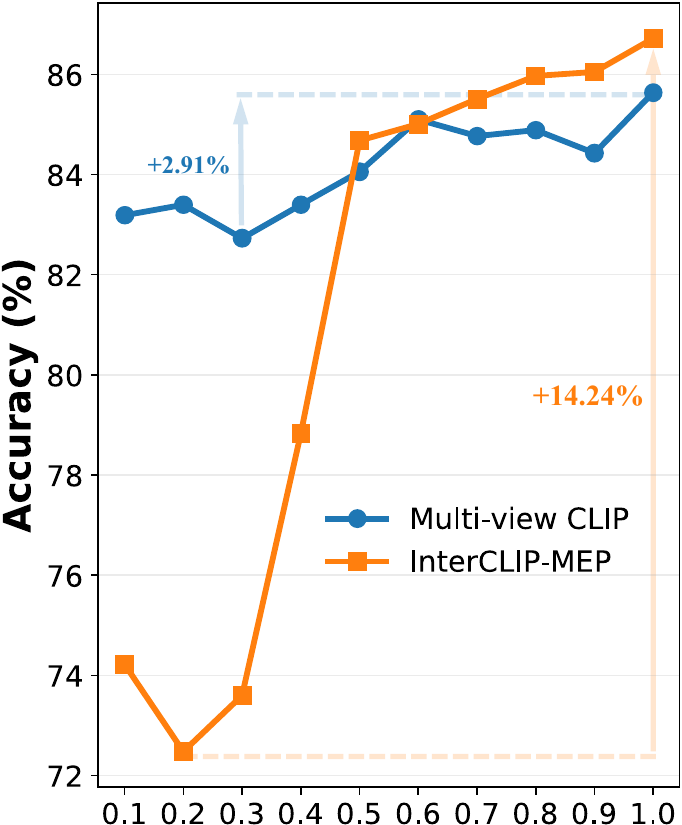}
\caption{
Accuracy on MMSD2.0 with different training ratios.
}
\label{fig:low-res-exp}

\centering
\includegraphics[width=0.75\linewidth]{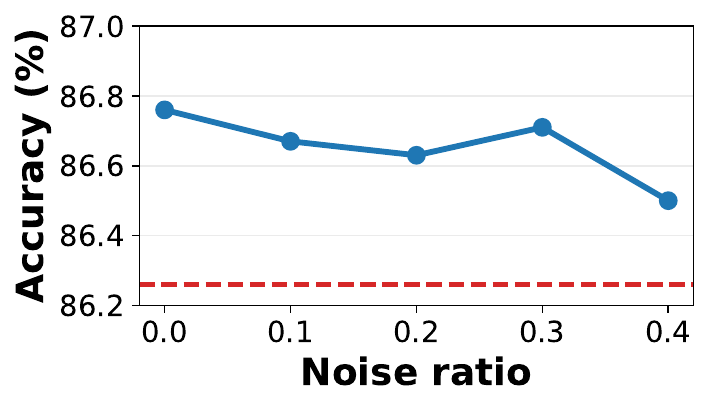}
\caption{
Accuracy under different levels of pseudo-label noise.
The dashed line indicates the result without MEP.
}
\label{fig:infer-error-prop-exp}

\end{minipage}\hfill
\begin{minipage}[c]{0.5\textwidth}
\centering
\includegraphics[width=0.88\linewidth]{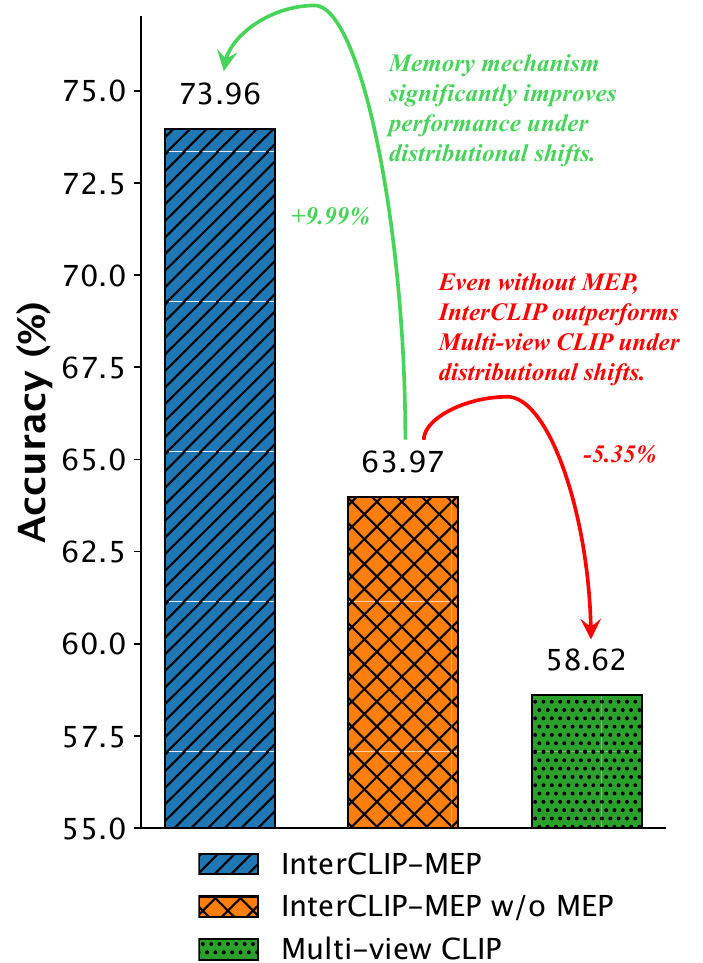}
\caption{
Evaluation on a mixed MMSD2.0–DocMSU test set illustrates that interactive representations and the memory mechanism jointly enhance robustness under distributional shifts.
}
\label{fig:infer-dist-shift-exp}
\end{minipage}
\end{figure}

To assess the robustness of InterCLIP-MEP under low-resource conditions, we followed the experimental protocol of Multi-view CLIP~\cite{qin2023mmsd2} and conducted experiments on the MMSD2.0 dataset.
Specifically, we subsampled the training set by selecting both positive and negative samples at equal intervals of 10\%, and employed the hyperparameter configuration that yielded the best performance in Table~\ref{tab:main-results}.
The models were then evaluated on the test set, and the results are presented in Figure~\ref{fig:low-res-exp}.

As shown in the figure, when 50\% or more of the training data is available, InterCLIP-MEP achieves a clear advantage over Multi-view CLIP.
This indicates that our framework is capable of leveraging moderately limited supervision to reconstruct a more effective multi-modal representation space.
Moreover, while our method performs slightly worse than Multi-view CLIP when training with less than 50\% of the data, it exhibits an emergent performance gain once this threshold is reached. Specifically, the accuracy of InterCLIP-MEP increases by 14.24\% from its lowest to highest point, whereas Multi-view CLIP only improves by 2.91\%.

We attribute this phenomenon to the distinct training strategies.
Multi-view CLIP relies on original CLIP representations, which provide relatively strong performance even with limited supervision but yield diminishing returns as more data becomes available.
In contrast, InterCLIP-MEP employs lightweight fine-tuning to restructure the representation space of CLIP, making it more aligned with the specific requirements of multi-modal sarcasm detection.
This process requires a sufficient amount of data to stabilize the new cross-modal alignment, but once the data threshold is met, the framework benefits substantially more from additional training samples.

\subsection{Error Propagation Robustness Study} \label{sec:err-prop-exp}
To further assess the robustness of the proposed memory mechanism against error propagation, we designed a stress test by randomly flipping a certain proportion of pseudo-labels generated by the classification module. This operation introduces deliberately corrupted low-entropy predictions into the memory, thereby simulating potential error accumulation during inference. For consistency, all hyperparameters in this experiment follow the optimal settings reported in Table~\ref{tab:main-results}.

The results are shown in Figure~\ref{fig:infer-error-prop-exp}. Even when up to 40\% of pseudo-labels are randomly flipped, the accuracy of InterCLIP-MEP remains consistently above the baseline system without memory (86.26\%). The observed variations are minimal, indicating that the memory-enhanced predictor maintains stable performance under noisy pseudo-labels.
These findings demonstrate that the memory-enhanced predictor does not suffer from cascading error amplification. Instead, it maintains stable performance under noisy pseudo-labels, confirming that the dynamic update and entropy-based filtering strategy effectively suppress the influence of mislabeled entries.

\subsection{Robustness under Distributional Shift} \label{sec:dist-shift-exp}
To rigorously examine robustness under distributional shifts, we compared InterCLIP-MEP, its variant without the memory mechanism (w/o MEP), and the state-of-the-art Multi-view CLIP baseline.
All models were trained on the MMSD2.0~\cite{qin2023mmsd2} training set and evaluated on a mixed testing set constructed by shuffling the MMSD2.0 and DocMSU~\cite{du2024docmsu} test sets. This heterogeneous evaluation setting introduces significant domain variation beyond the training distribution and provides a stringent test of model stability.

As shown in Figure~\ref{fig:infer-dist-shift-exp}, InterCLIP-MEP attains the highest accuracy of 73.96\%, substantially outperforming its variant without the memory mechanism (63.97\%) and Multi-view CLIP (58.62\%).
The nearly 10\% improvement over the w/o MEP variant demonstrates that the dynamic memory mechanism significantly mitigates performance degradation by suppressing error accumulation and leveraging historical knowledge during inference.
Moreover, even without the memory component, InterCLIP itself still surpasses Multi-view CLIP by more than 5\%, highlighting the effectiveness of the interactive representation design.
These results confirm that InterCLIP-MEP achieves superior stability and adaptability when faced with distributional shifts.

\section{Limitations and Future Work}\label{sec:limitation-and-future-works}
While InterCLIP-MEP demonstrates strong robustness and consistent improvements across multiple evaluations, several limitations remain that highlight promising directions for future research.

\paragraph{Limitations.}
Despite its effectiveness, InterCLIP-MEP can be sensitive to \textit{memory noise} when misleading or weakly relevant representations are retrieved, which may bias the inference process and occasionally override correct predictions.
Moreover, the framework still struggles with \textit{highly implicit or context-dependent sarcasm}, especially when such cues rely on external commonsense or situational background that is not explicitly encoded in the available modalities.
In addition, the current design assumes a relatively static memory and homogeneous data distribution, which may limit adaptability when facing unseen domains or evolving data patterns.

\paragraph{Future Work.}
Future studies may explore context-aware memory updating strategies that dynamically adjust to semantic drift and data uncertainty.
Integrating external commonsense or world knowledge could further enhance the model’s ability to capture subtle pragmatic cues and hidden irony.
Another potential direction is to introduce memory regularization or selective forgetting mechanisms to mitigate the effect of noisy retrievals and reduce error propagation under domain shifts.
Finally, extending the proposed memory-augmented paradigm to other multimodal reasoning tasks, such as emotion understanding or visual entailment, could provide a broader understanding of how long-term memory interacts with perception and language in open-world scenarios.

\section{Conclusion}\label{sec:conclusion}
This paper introduces InterCLIP-MEP, an innovative framework designed to address the critical challenges of modeling subtle text-image interactions and mitigating prediction uncertainty in multi-modal sarcasm detection.
The proposed framework incorporates Interactive CLIP (InterCLIP), which integrates cross-modal information directly into text and image encoders.
This enhances the model's ability to interpret multi-modal sarcasm cues.
Furthermore, a Memory-Enhanced Predictor (MEP) is developed to dynamically utilize knowledge from historical samples, resulting in a more robust and adaptive inference process.
Extensive evaluations conducted on the MMSD, MMSD2.0, and DocMSU benchmarks demonstrate that InterCLIP-MEP achieves state-of-the-art performance while substantially reducing computational overhead.
By requiring fewer trainable parameters and less GPU memory, InterCLIP-MEP offers a lightweight, efficient, and scalable solution, setting a new benchmark for multi-modal sarcasm detection.

\section*{Acknowledgements}
The research presented in this paper was funded by the Anhui Provincial Natural Science Foundation under Grant No. 2308085MF220.
It also received additional support from the Anhui University Natural Science Foundation under Grant No. 2023AH050914.

\bibliographystyle{ACM-Reference-Format}
\bibliography{references}

@article{JoshiBC17,
  title        = {Automatic Sarcasm Detection: {A} Survey},
  author       = {Aditya Joshi and Pushpak Bhattacharyya and Mark James Carman},
  year         = 2017,
  journal      = {{ACM} Comput. Surv.},
  volume       = 50,
  number       = 5,
  pages        = {73:1--73:22}
}

@inproceedings{JoshiTPBC16,
  title        = {Are Word Embedding-based Features Useful for Sarcasm Detection?},
  author       = {Aditya Joshi and Vaibhav Tripathi and Kevin Patel and Pushpak Bhattacharyya and Mark James Carman},
  year         = 2016,
  booktitle    = {Proceedings of the 2016 Conference on Empirical Methods in Natural Language Processing, {EMNLP} 2016, Austin, Texas, USA, November 1-4},
  pages        = {1006--1011}
}

@article{WangYJMXL24,
  title        = {Cross-modal incongruity aligning and collaborating for multi-modal sarcasm detection},
  author       = {Jie Wang and Yan Yang and Yongquan Jiang and Minbo Ma and Zhuyang Xie and Tianrui Li},
  year         = 2024,
  journal      = {Inf. Fusion},
  volume       = 103,
  pages        = 102132
}

@inproceedings{JingSOJN23,
  title        = {Multi-source Semantic Graph-based Multimodal Sarcasm Explanation Generation},
  author       = {Liqiang Jing and Xuemeng Song and Kun Ouyang and Mengzhao Jia and Liqiang Nie},
  year         = 2023,
  booktitle    = {Proceedings of the 61st Annual Meeting of the Association for Computational Linguistics (Volume 1: Long Papers), Toronto, Canada, July 9-14},
  pages        = {11349--11361}
}

@article{LiuZS24,
  title        = {A Quantum Probability Driven Framework for Joint Multi-Modal Sarcasm, Sentiment and Emotion Analysis},
  author       = {Yaochen Liu and Yazhou Zhang and Dawei Song},
  year         = 2024,
  journal      = {{IEEE} Trans. Affect. Comput.},
  volume       = 15,
  number       = 1,
  pages        = {326--341}
}

@article{ZhuangLZGZM25,
  title        = {DyCR-Net: {A} dynamic context-aware routing network for multi-modal sarcasm detection in conversation},
  author       = {Xingjie Zhuang and Zhixin Li and Fengling Zhou and Jingliang Gu and Canlong Zhang and Huifang Ma},
  year         = 2025,
  journal      = {Knowl. Based Syst.},
  volume       = 310,
  pages        = 113029
}

@inproceedings{amir2016modelling,
  title        = {Modelling Context with User Embeddings for Sarcasm Detection in Social Media},
  author       = {Amir, Silvio  and Wallace, Byron C.  and Lyu, Hao  and Carvalho, Paula  and Silva, M{\'a}rio J.},
  year         = 2016,
  month        = aug,
  booktitle    = {Proceedings of the 20th {SIGNLL} Conference on Computational Natural Language Learning},
  address      = {Berlin, Germany},
  pages        = {167--177}
}

@article{Baddeley2000TheEB,
  title        = {The episodic buffer: a new component of working memory?},
  author       = {Alan D. Baddeley},
  year         = 2000,
  journal      = {Trends in Cognitive Sciences},
  volume       = 4,
  pages        = {417--423}
}

@inproceedings{baziotis2018ntua,
  title        = {{NTUA}-{SLP} at {S}em{E}val-2018 Task 3: Tracking Ironic Tweets using Ensembles of Word and Character Level Attentive {RNN}s},
  author       = {Baziotis, Christos  and Nikolaos, Athanasiou  and Papalampidi, Pinelopi  and Kolovou, Athanasia  and Paraskevopoulos, Georgios  and Ellinas, Nikolaos  and Potamianos, Alexandros},
  year         = 2018,
  month        = jun,
  booktitle    = {Proceedings of the 12th International Workshop on Semantic Evaluation},
  pages        = {613--621}
}

@inproceedings{bouazizi2015sarcasm,
  title        = {Sarcasm detection in twitter:" all your products are incredibly amazing!!!"-are they really?},
  author       = {Bouazizi, Mondher and Ohtsuki, Tomoaki},
  year         = 2015,
  booktitle    = {2015 IEEE global communications conference (GLOBECOM)},
  pages        = {1--6}
}

@inproceedings{Cai2019MultiModalSD,
  title        = {Multi-Modal Sarcasm Detection in Twitter with Hierarchical Fusion Model},
  author       = {Yitao Cai and Huiyu Cai and Xiaojun Wan},
  year         = 2019,
  booktitle    = {Proceedings of the 57th Conference of the Association for Computational Linguistics, {ACL} 2019, Florence, Italy, July 28- August 2, 2019, Volume 1: Long Papers},
  pages        = {2506--2515}
}

@inproceedings{devlin2019bert,
  title        = {{BERT:} Pre-training of Deep Bidirectional Transformers for Language Understanding},
  author       = {Jacob Devlin and Ming{-}Wei Chang and Kenton Lee and Kristina Toutanova},
  year         = 2019,
  booktitle    = {Proceedings of the 2019 Conference of the North American Chapter of the Association for Computational Linguistics: Human Language Technologies, {NAACL-HLT} 2019, Minneapolis, MN, USA, June 2-7, 2019, Volume 1 (Long and Short Papers)},
  pages        = {4171--4186}
}

@inproceedings{dosovitskiy2020image,
  title        = {An Image is Worth 16x16 Words: Transformers for Image Recognition at Scale},
  author       = {Alexey Dosovitskiy and Lucas Beyer and Alexander Kolesnikov and Dirk Weissenborn and Xiaohua Zhai and Thomas Unterthiner and Mostafa Dehghani and Matthias Minderer and Georg Heigold and Sylvain Gelly and Jakob Uszkoreit and Neil Houlsby},
  year         = 2021,
  booktitle    = {9th International Conference on Learning Representations, {ICLR} 2021, Virtual Event, Austria, May 3-7, 2021},
  pages        = {1--21}
}

@inproceedings{ganz2024question,
  title        = {Question Aware Vision Transformer for Multimodal Reasoning},
  author       = {Ganz, Roy and Kittenplon, Yair and Aberdam, Aviad and Ben Avraham, Elad and Nuriel, Oren and Mazor, Shai and Litman, Ron},
  year         = 2024,
  month        = {June},
  booktitle    = {Proceedings of the IEEE/CVF Conference on Computer Vision and Pattern Recognition (CVPR)},
  pages        = {13861--13871}
}

@article{Gibbs1991PsychologicalAO,
  title        = {Psychological aspects of irony understanding},
  author       = {Raymond W. Gibbs and Jennifer O'Brien},
  year         = 1991,
  journal      = {Journal of Pragmatics},
  volume       = 16,
  number       = 6,
  pages        = {523--530}
}

@inproceedings{zhou-etal-2016-attention,
  title        = {Attention-Based Bidirectional Long Short-Term Memory Networks for Relation Classification},
  author       = {Zhou, Peng  and Shi, Wei  and Tian, Jun  and Qi, Zhenyu  and Li, Bingchen  and Hao, Hongwei  and Xu, Bo},
  year         = 2016,
  month        = aug,
  booktitle    = {Proceedings of the 54th Annual Meeting of the Association for Computational Linguistics (Volume 2: Short Papers)},
  pages        = {207--212}
}

@inproceedings{he2015deep,
  title        = {Deep Residual Learning for Image Recognition},
  author       = {He, Kaiming and Zhang, Xiangyu and Ren, Shaoqing and Sun, Jian},
  year         = 2016,
  booktitle    = {2016 IEEE Conference on Computer Vision and Pattern Recognition (CVPR)},
  pages        = {770--778}
}

@inproceedings{edward2022lora,
  title        = {Lo{RA}: Low-Rank Adaptation of Large Language Models},
  author       = {Edward J Hu and Yelong Shen and Phillip Wallis and Zeyuan Allen-Zhu and Yuanzhi Li and Shean Wang and Lu Wang and Weizhu Chen},
  year         = 2022,
  booktitle    = {International Conference on Learning Representations},
  pages        = {1--13}
}

@inproceedings{kim-2014-convolutional,
  title        = {Convolutional Neural Networks for Sentence Classification},
  author       = {Kim, Yoon},
  year         = 2014,
  month        = oct,
  booktitle    = {Proceedings of the 2014 Conference on Empirical Methods in Natural Language Processing ({EMNLP})},
  pages        = {1746--1751}
}

@article{li2022adapting,
  title        = {Adapting {CLIP} For Phrase Localization Without Further Training},
  author       = {Jiahao Li and Greg Shakhnarovich and Raymond A. Yeh},
  year         = 2022,
  journal      = {CoRR},
  volume       = {abs/2204.03647},
  doi          = {10.48550/ARXIV.2204.03647},
  url          = {https://doi.org/10.48550/arXiv.2204.03647},
  eprinttype   = {arXiv},
  eprint       = {2204.03647}
}

@inproceedings{liang2021multi,
  title        = {Multi-modal sarcasm detection with interactive in-modal and cross-modal graphs},
  author       = {Liang, Bin and Lou, Chenwei and Li, Xiang and Gui, Lin and Yang, Min and Xu, Ruifeng},
  year         = 2021,
  booktitle    = {Proceedings of the 29th ACM international conference on multimedia},
  pages        = {4707--4715}
}

@inproceedings{Liang2022MultiModalSD,
  title        = {Multi-Modal Sarcasm Detection via Cross-Modal Graph Convolutional Network},
  author       = {Bin Liang and Chenwei Lou and Xiang Li and Min Yang and Lin Gui and Yulan He and Wenjie Pei and Ruifeng Xu},
  year         = 2022,
  booktitle    = {Proceedings of the 60th Annual Meeting of the Association for Computational Linguistics (Volume 1: Long Papers), {ACL} 2022, Dublin, Ireland, May 22-27},
  pages        = {1767--1777}
}

@inproceedings{Liang2023Open,
  title        = {Open-Vocabulary Semantic Segmentation With Mask-Adapted CLIP},
  author       = {Liang, Feng and Wu, Bichen and Dai, Xiaoliang and Li, Kunpeng and Zhao, Yinan and Zhang, Hang and Zhang, Peizhao and Vajda, Peter and Marculescu, Diana},
  year         = 2023,
  month        = {June},
  booktitle    = {Proceedings of the IEEE/CVF Conference on Computer Vision and Pattern Recognition (CVPR)},
  pages        = {7061--7070}
}

@inproceedings{Liu2022TowardsMS,
  title        = {Towards Multi-Modal Sarcasm Detection via Hierarchical Congruity Modeling with Knowledge Enhancement},
  author       = {Hui Liu and Wenya Wang and Haoliang Li},
  year         = 2022,
  booktitle    = {Proceedings of the 2022 Conference on Empirical Methods in Natural Language Processing, {EMNLP} 2022, Abu Dhabi, United Arab Emirates, December 7-11},
  pages        = {4995--5006}
}

@article{liu2019roberta,
  title        = {RoBERTa: {A} Robustly Optimized {BERT} Pretraining Approach},
  author       = {Yinhan Liu and Myle Ott and Naman Goyal and Jingfei Du and Mandar Joshi and Danqi Chen and Omer Levy and Mike Lewis and Luke Zettlemoyer and Veselin Stoyanov},
  year         = 2019,
  journal      = {CoRR},
  volume       = {abs/1907.11692},
  url          = {http://arxiv.org/abs/1907.11692},
  eprinttype   = {arXiv}
}

@inproceedings{pan-etal-2020-modeling,
  title        = {Modeling Intra and Inter-modality Incongruity for Multi-Modal Sarcasm Detection},
  author       = {Hongliang Pan and Zheng Lin and Peng Fu and Yatao Qi and Weiping Wang},
  year         = 2020,
  booktitle    = {Findings of the Association for Computational Linguistics: {EMNLP} 2020, Online Event, 16-20 November 2020},
  pages        = {1383--1392}
}

@article{pang2008opinion,
  title        = {Opinion mining and sentiment analysis},
  author       = {Pang, Bo and Lee, Lillian and others},
  year         = 2008,
  journal      = {Foundations and Trends in information retrieval},
  publisher    = {Now Publishers, Inc.},
  volume       = 2,
  number       = {1--2},
  pages        = {1--135}
}

@inproceedings{qin2023mmsd2,
  title        = {{MMSD}2.0: Towards a Reliable Multi-modal Sarcasm Detection System},
  author       = {Qin, Libo  and Huang, Shijue  and Chen, Qiguang  and Cai, Chenran  and Zhang, Yudi  and Liang, Bin  and Che, Wanxiang  and Xu, Ruifeng},
  year         = 2023,
  month        = jul,
  booktitle    = {Findings of the Association for Computational Linguistics: ACL 2023},
  address      = {Toronto, Canada},
  pages        = {10834--10845}
}

@inproceedings{radford2021learning,
  title        = {Learning transferable visual models from natural language supervision},
  author       = {Radford, Alec and Kim, Jong Wook and Hallacy, Chris and Ramesh, Aditya and Goh, Gabriel and Agarwal, Sandhini and Sastry, Girish and Askell, Amanda and Mishkin, Pamela and Clark, Jack and others},
  year         = 2021,
  booktitle    = {International conference on machine learning},
  pages        = {8748--8763}
}

@inproceedings{schifanella2016detecting,
  title        = {Detecting sarcasm in multimodal social platforms},
  author       = {Schifanella, Rossano and De Juan, Paloma and Tetreault, Joel and Cao, Liangliang},
  year         = 2016,
  booktitle    = {Proceedings of the 24th ACM international conference on Multimedia},
  pages        = {1136--1145}
}

@article{stokes2015activity,
  title        = {‘Activity-silent’working memory in prefrontal cortex: a dynamic coding framework},
  author       = {Stokes, Mark G},
  year         = 2015,
  journal      = {Trends in cognitive sciences},
  volume       = 19,
  number       = 7,
  pages        = {394--405}
}

@inproceedings{sukhbaatar2015end,
  title        = {End-To-End Memory Networks},
  author       = {Sainbayar Sukhbaatar and Arthur Szlam and Jason Weston and Rob Fergus},
  year         = 2015,
  booktitle    = {Advances in Neural Information Processing Systems 28: Annual Conference on Neural Information Processing Systems 2015, December 7-12, 2015, Montreal, Quebec, Canada},
  pages        = {2440--2448}
}

@inproceedings{Tian2023DynamicRT,
  title        = {Dynamic Routing Transformer Network for Multimodal Sarcasm Detection},
  author       = {Yuan Tian and Nan Xu and Ruike Zhang and Wenji Mao},
  year         = 2023,
  booktitle    = {Proceedings of the 61st Annual Meeting of the Association for Computational Linguistics (Volume 1: Long Papers), {ACL} 2023, Toronto, Canada, July 9-14},
  pages        = {2468--2480}
}

@inproceedings{tsur2010icwsm,
  title        = {{ICWSM} - {A} Great Catchy Name: Semi-Supervised Recognition of Sarcastic Sentences in Online Product Reviews},
  author       = {Oren Tsur and Dmitry Davidov and Ari Rappoport},
  year         = 2010,
  booktitle    = {Proceedings of the Fourth International Conference on Weblogs and Social Media, {ICWSM} 2010, Washington, DC, USA, May 23-26},
  pages        = {1--8}
}

@inproceedings{wang2023seeing,
  title        = {Seeing in Flowing: Adapting {CLIP} for Action Recognition with Motion Prompts Learning},
  author       = {Qiang Wang and Junlong Du and Ke Yan and Shouhong Ding},
  year         = 2023,
  booktitle    = {Proceedings of the 31st {ACM} International Conference on Multimedia, {MM} 2023, Ottawa, ON, Canada, 29 October 2023- 3 November},
  pages        = {5339--5347}
}

@inproceedings{wei2024g2sam,
  title        = {G{\^{}}2SAM: Graph-Based Global Semantic Awareness Method for Multimodal Sarcasm Detection},
  author       = {Yiwei Wei and Shaozu Yuan and Hengyang Zhou and Longbiao Wang and Zhiling Yan and Ruosong Yang and Meng Chen},
  year         = 2024,
  booktitle    = {Thirty-Eighth {AAAI} Conference on Artificial Intelligence, {AAAI} 2024, Thirty-Sixth Conference on Innovative Applications of Artificial Intelligence, {IAAI} 2024, Fourteenth Symposium on Educational Advances in Artificial Intelligence, {EAAI} 2014, February 20-27, 2024, Vancouver, Canada},
  pages        = {9151--9159}
}

@inproceedings{wen2023dip,
  title        = {Dip: Dual incongruity perceiving network for sarcasm detection},
  author       = {Wen, Changsong and Jia, Guoli and Yang, Jufeng},
  year         = 2023,
  booktitle    = {Proceedings of the IEEE/CVF Conference on Computer Vision and Pattern Recognition},
  pages        = {2540--2550}
}

@article{weston2014memory,
  title        = {Memory Networks},
  author       = {J. Weston and S. Chopra and Antoine Bordes},
  year         = 2014,
  journal      = {International Conference on Learning Representations},
  pages        = {2540--2550}
}

@inproceedings{wu2018unsupervised,
  title        = {Unsupervised feature learning via non-parametric instance discrimination},
  author       = {Wu, Zhirong and Xiong, Yuanjun and Yu, Stella X and Lin, Dahua},
  year         = 2018,
  booktitle    = {Proceedings of the IEEE conference on computer vision and pattern recognition},
  pages        = {3733--3742}
}

@inproceedings{xiong2019sarcasm,
  title        = {Sarcasm Detection with Self-matching Networks and Low-rank Bilinear Pooling},
  author       = {Tao Xiong and Peiran Zhang and Hongbo Zhu and Yihui Yang},
  year         = 2019,
  booktitle    = {The World Wide Web Conference, {WWW} 2019, San Francisco, CA, USA, May 13-17},
  pages        = {2115--2124}
}

@inproceedings{xu2020reasoning,
  title        = {Reasoning with Multimodal Sarcastic Tweets via Modeling Cross-Modality Contrast and Semantic Association},
  author       = {Nan Xu and Zhixiong Zeng and Wenji Mao},
  year         = 2020,
  booktitle    = {Proceedings of the 58th Annual Meeting of the Association for Computational Linguistics, {ACL} 2020, Online, July 5-10},
  pages        = {3777--3786}
}

@inproceedings{zhang2024dual,
  title        = {Dual Memory Networks: {A} Versatile Adaptation Approach for Vision-Language Models},
  author       = {Yabin Zhang and Wenjie Zhu and Hui Tang and Zhiyuan Ma and Kaiyang Zhou and Lei Zhang},
  year         = 2024,
  booktitle    = {{IEEE/CVF} Conference on Computer Vision and Pattern Recognition, {CVPR} 2024, Seattle, WA, USA, June 16-22},
  pages        = {28718--28728}
}

@inproceedings{Selvaraju2017GradCAM,
  title        = {Grad-CAM: Visual Explanations from Deep Networks via Gradient-Based Localization},
  author       = {Selvaraju, Ramprasaath R. and Cogswell, Michael and Das, Abhishek and Vedantam, Ramakrishna and Parikh, Devi and Batra, Dhruv},
  year         = 2017,
  month        = {Oct},
  booktitle    = {2017 IEEE International Conference on Computer Vision (ICCV)},
  pages        = {618--626}
}

@inproceedings{wolf-etal-2020-transformers,
  title        = {Transformers: State-of-the-Art Natural Language Processing},
  author       = {Thomas Wolf and Lysandre Debut and Victor Sanh and Julien Chaumond and Clement Delangue and Anthony Moi and Pierric Cistac and Tim Rault and Rémi Louf and Morgan Funtowicz and Joe Davison and Sam Shleifer and Patrick von Platen and Clara Ma and Yacine Jernite and Julien Plu and Canwen Xu and Teven Le Scao and Sylvain Gugger and Mariama Drame and Quentin Lhoest and Alexander M. Rush},
  year         = 2020,
  month        = oct,
  booktitle    = {Proceedings of the 2020 Conference on Empirical Methods in Natural Language Processing: System Demonstrations},
  pages        = {38--45}
}

@inproceedings{loshchilov2017decoupled,
  title        = {Decoupled Weight Decay Regularization},
  author       = {Ilya Loshchilov and Frank Hutter},
  year         = 2019,
  booktitle    = {7th International Conference on Learning Representations, {ICLR} 2019, New Orleans, LA, USA, May 6-9},
  pages        = {1--18}
}

@inproceedings{du2024docmsu,
  title        = {DocMSU: A Comprehensive Benchmark for Document-Level Multimodal Sarcasm Understanding},
  author       = {Du, Hang and Nan, Guoshun and Zhang, Sicheng and Xie, Binzhu and Xu, Junrui and Fan, Hehe and Cui, Qimei and Tao, Xiaofeng and Jiang, Xudong},
  year         = 2024,
  booktitle    = {Proceedings of the AAAI Conference on Artificial Intelligence},
  volume       = 38,
  pages        = {17933--17941}
}

@inproceedings{liu2021swin,
  title        = {Swin transformer: Hierarchical vision transformer using shifted windows},
  author       = {Liu, Ze and Lin, Yutong and Cao, Yue and Hu, Han and Wei, Yixuan and Zhang, Zheng and Lin, Stephen and Guo, Baining},
  year         = 2021,
  booktitle    = {Proceedings of the IEEE/CVF international conference on computer vision},
  pages        = {10012--10022}
}

@inproceedings{kim2021vilt,
  title        = {ViLT: Vision-and-Language Transformer Without Convolution or Region Supervision},
  author       = {Wonjae Kim and Bokyung Son and Ildoo Kim},
  year         = 2021,
  booktitle    = {Proceedings of the 38th International Conference on Machine Learning, {ICML} 2021, 18-24 July 2021, Virtual Event},
  volume       = 139,
  pages        = {5583--5594}
}

@inproceedings{tang-etal-2024-leveraging,
  title        = {Leveraging Generative Large Language Models with Visual Instruction and Demonstration Retrieval for Multimodal Sarcasm Detection},
  author       = {Tang, Binghao  and Lin, Boda  and Yan, Haolong  and Li, Si},
  year         = 2024,
  month        = jun,
  booktitle    = {Proceedings of the 2024 Conference of the North American Chapter of the Association for Computational Linguistics: Human Language Technologies (Volume 1: Long Papers)},
  address      = {Mexico City, Mexico},
  pages        = {1732--1742}
}

@article{Wang2025AdaSFFuse,
  title        = {Task-Generalized Adaptive Cross-Domain Learning for Multimodal Image Fusion},
  author       = {Mengyu Wang and Zhenyu Liu and Kun Li and Yu Wang and Yuwei Wang and Yanyan Wei and Fei Wang},
  year         = 2025,
  journal      = {arXiv preprint arXiv:2508.15505}
}

@article{anderson1991reflections,
  title        = {Reflections of the environment in memory},
  author       = {Anderson, John R and Schooler, Lael J},
  year         = 1991,
  journal      = {Psychological science},
  publisher    = {SAGE Publications Sage CA: Los Angeles, CA},
  volume       = 2,
  number       = 6,
  pages        = {396--408}
}

@article{settles2009active,
  title        = {Active learning literature survey},
  author       = {Settles, Burr},
  year         = 2009,
  publisher    = {University of Wisconsin-Madison Department of Computer Sciences}
}

@article{toneva2018empirical,
  title        = {An empirical study of example forgetting during deep neural network learning},
  author       = {Toneva, Mariya and Sordoni, Alessandro and Combes, Remi Tachet des and Trischler, Adam and Bengio, Yoshua and Gordon, Geoffrey J},
  year         = 2018,
  journal      = {arXiv preprint arXiv:1812.05159}
}

@article{attardo2000irony,
  title        = {Irony as relevant inappropriateness},
  author       = {Attardo, Salvatore},
  year         = 2000,
  journal      = {Journal of pragmatics},
  publisher    = {Elsevier},
  volume       = 32,
  number       = 6,
  pages        = {793--826}
}

@article{gibbs1986psycholinguistics,
  title        = {On the psycholinguistics of sarcasm.},
  author       = {Gibbs, Raymond W},
  year         = 1986,
  journal      = {Journal of experimental psychology: general},
  publisher    = {American Psychological Association},
  volume       = 115,
  number       = 1,
  pages        = 3
}

@book{gibbs2007irony,
  title        = {Irony in language and thought: A cognitive science reader},
  author       = {Gibbs, Raymond W and Colston, Herbert L},
  year         = 2007,
  publisher    = {Psychology Press}
}

@inproceedings{bamman2015contextualized,
  title        = {Contextualized sarcasm detection on twitter},
  author       = {Bamman, David and Smith, Noah},
  year         = 2015,
  booktitle    = {proceedings of the international AAAI conference on web and social media},
  volume       = 9,
  number       = 1,
  pages        = {574--577}
}

@article{zhu2020label,
  title        = {Label independent memory for semi-supervised few-shot video classification},
  author       = {Zhu, Linchao and Yang, Yi},
  year         = 2020,
  journal      = {IEEE Transactions on Pattern Analysis and Machine Intelligence},
  publisher    = {IEEE},
  volume       = 44,
  number       = 1,
  pages        = {273--285}
}

@inproceedings{10.24963/ijcai.2024/887,
  title        = {A survey of multimodal sarcasm detection},
  author       = {Farabi, Shafkat and Ranasinghe, Tharindu and Kanojia, Diptesh and Kong, Yu and Zampieri, Marcos},
  year         = 2024,
  booktitle    = {Proceedings of the Thirty-Third International Joint Conference on Artificial Intelligence},
  location     = {Jeju, Korea},
  series       = {IJCAI '24},
  doi          = {10.24963/ijcai.2024/887},
  isbn         = {978-1-956792-04-1},
  url          = {https://doi.org/10.24963/ijcai.2024/887},
  articleno    = 887,
  numpages     = 9
}

@inproceedings{ding2022multi,
  title        = {Multi-modal sarcasm detection with prompt-tuning},
  author       = {Ding, Daijun and Huang, Hu and Zhang, Bowen and Peng, Cheng and Li, Yangyang and Fu, Xianghua and Jing, Liwen},
  year         = 2022,
  booktitle    = {2022 6th Asian Conference on Artificial Intelligence Technology (ACAIT)},
  pages        = {1--8},
  organization = {IEEE}
}

@inproceedings{wallace-etal-2014-humans,
  title        = {Humans Require Context to Infer Ironic Intent (so Computers Probably do, too)},
  author       = {Wallace, Byron C.  and Choe, Do Kook  and Kertz, Laura  and Charniak, Eugene},
  year         = 2014,
  month        = jun,
  booktitle    = {Proceedings of the 52nd Annual Meeting of the Association for Computational Linguistics (Volume 2: Short Papers)},
  publisher    = {Association for Computational Linguistics},
  address      = {Baltimore, Maryland},
  pages        = {512--516},
  doi          = {10.3115/v1/P14-2084},
  url          = {https://aclanthology.org/P14-2084/},
  editor       = {Toutanova, Kristina  and Wu, Hua}
}

@article{10477507,
  title        = {Fusion and Discrimination: A Multimodal Graph Contrastive Learning Framework for Multimodal Sarcasm Detection},
  author       = {Liang, Bin and Gui, Lin and He, Yulan and Cambria, Erik and Xu, Ruifeng},
  year         = 2024,
  journal      = {IEEE Transactions on Affective Computing},
  volume       = 15,
  number       = 4,
  pages        = {1874--1888},
  doi          = {10.1109/TAFFC.2024.3380375}
}

@article{zhuang2025multi,
  title        = {Multi-Modal Sarcasm Detection via Knowledge-Aware Focused Graph Convolutional Networks},
  author       = {Zhuang, Xingjie and Zhou, Fengling and Li, Zhixin},
  year         = 2025,
  journal      = {ACM Transactions on Multimedia Computing, Communications and Applications},
  publisher    = {ACM New York, NY},
  volume       = 21,
  number       = 5,
  pages        = {1--22}
}


\end{document}